\documentclass{vldb}
\usepackage{graphicx}
\usepackage{balance}  

\title{Dynamic Parameter Allocation in Parameter Servers}

\numberofauthors{1} 

\author{
  \alignauthor
  Alexander Renz-Wieland{\small $~^{1}$}, Rainer Gemulla{\small $~^{2}$}, Steffen Zeuch{\small $~^{1,3}$}, Volker Markl{\small $~^{1,3}$}\\
  \affaddr{\vspace{2mm}\mbox{$^1$Technische Universit\"at Berlin, $^2$Universit\"at Mannheim, $^3$German Research
      Center for Artificial Intelligence}}\\
  \affaddr{\vspace{1mm}$^1$firstname.lastname@tu-berlin.de, $^2$rgemulla@uni-mannheim.de, $^3$firstname.lastname@dfki.de}
}

\vldbTitle{Dynamic Parameter Allocation in Parameter Servers}
\vldbAuthors{Alexander Renz-Wieland, Rainer Gemulla, Steffen Zeuch, Volker Markl}
\vldbDOI{https://doi.org/10.14778/3407790.3407796}
\vldbVolume{13}
\vldbNumber{11}
\vldbYear{2020}

\usepackage[normalem]{ulem}
\usepackage{amsmath}
\usepackage{amssymb}
\usepackage{hyperref}
\usepackage{tikz}
\usepackage{comment}
\usetikzlibrary{shapes,snakes,shapes.geometric,calc,arrows,matrix,shapes.arrows}
\usepackage{siunitx}
\usepackage{booktabs}
\usepackage{threeparttable}
\usepackage{subcaption}
\usepackage{array}
\usepackage{amsxtra}
\usepackage{textgreek}
\usepackage[labelfont=bf]{caption}

\newcommand{\eat}[1]{} 

\newcommand{\td}[2]{\if*#1\else^{#1}\fi\if*#2\else_{#2}\fi} 

\newcommand\join\Join 

\DeclareSymbolFont{txsymbolsC}{U}{txsyc}{m}{n}
\SetSymbolFont{txsymbolsC}{bold}{U}{txsyc}{bx}{n}
\DeclareFontSubstitution{U}{txsyc}{m}{n}
\DeclareMathSymbol{\ljoin}{\mathrel}{txsymbolsC}{88}
\DeclareMathSymbol{\rjoin}{\mathrel}{txsymbolsC}{89}

\newsavebox\setminusbox
\newlength\setminuslen

\newcolumntype{C}{>{$\displaystyle}c<{$}} 
\newcolumntype{L}{>{$\displaystyle}l<{$}} 
\newcolumntype{R}{>{$\displaystyle}r<{$}} 
\newcolumntype{H}{>{\setbox0=\hbox\bgroup}c<{\egroup}@{}} 

\makeatletter
\renewcommand*\env@matrix[1][*\c@MaxMatrixCols c]{%
  \hskip -\arraycolsep
  \let\@ifnextchar\new@ifnextchar
  \array{#1}}
\makeatother

\newcommand{\B}[3]{B\if*#1\else_{#1}\fi(#2,#3)} 
\newcommand{\I}[3]{I\if*#1\else_{#1}\fi(#2,#3)} 

\makeatletter
\def\imod#1{\allowbreak\mkern10mu({\operator@font mod}\,\,#1)}
\makeatother

\newlength\hspaceoflen

\newcommand\vect[1]{{\boldsymbol{#1}}}
\newcommand\va{\vect{a}}
\newcommand\vb{\vect{b}}
\newcommand\vc{\vect{c}}
\newcommand\vd{\vect{d}}
\newcommand\ve{\vect{e}}
\newcommand\vf{\vect{f}}
\newcommand\vg{\vect{g}}
\newcommand\vh{\vect{h}}
\newcommand\vi{\vect{i}}
\newcommand\vj{\vect{j}}
\newcommand\vk{\vect{k}}
\newcommand\vl{\vect{l}}
\newcommand\vm{\vect{m}}
\newcommand\vn{\vect{n}}
\newcommand\vo{\vect{o}}
\newcommand\vp{\vect{p}}
\newcommand\vq{\vect{q}}
\newcommand\vr{\vect{r}}
\newcommand\vs{\vect{s}}
\newcommand\vt{\vect{t}}
\newcommand\vu{\vect{u}}
\newcommand\vv{\vect{v}}
\newcommand\vw{\vect{w}}
\newcommand\vx{\vect{x}}
\newcommand\vy{\vect{y}}
\newcommand\vz{\vect{z}}

\newcommand\mA{\vect{A}}
\newcommand\mB{\vect{B}}
\newcommand\mC{\vect{C}} 
\newcommand\mD{\vect{D}}
\newcommand\mE{\vect{E}}
\newcommand\mF{\vect{F}}
\newcommand\mG{\vect{G}}
\newcommand\mH{\vect{H}}
\newcommand\mI{\vect{I}}
\newcommand\mJ{\vect{J}}
\newcommand\mK{\vect{K}}
\newcommand\mL{\vect{L}}
\newcommand\mM{\vect{M}}
\newcommand\mN{\vect{N}} 
\newcommand\mO{\vect{O}}
\newcommand\mP{\vect{P}}
\newcommand\mQ{\vect{Q}} 
\newcommand\mR{\vect{R}} 
\newcommand\mS{\vect{S}}
\newcommand\mT{\vect{T}}
\newcommand\mU{\vect{U}}
\newcommand\mV{\vect{V}}
\newcommand\mW{\vect{W}}
\newcommand\mX{\vect{X}}
\newcommand\mY{\vect{Y}}
\newcommand\mZ{\vect{Z}}

\DeclareMathAlphabet{\mathcal}{OMS}{cmsy}{m}{n}

\DeclareMathAlphabet\mathbfcal{OMS}{cmsy}{b}{n}

\accentedsymbol\Abar{{\bar A}}
\accentedsymbol\Bbar{{\bar B}}
\accentedsymbol\Cbar{{\bar C}}
\accentedsymbol\Dbar{{\bar D}}
\accentedsymbol\Ebar{{\bar E}}
\accentedsymbol\Fbar{{\bar F}}
\accentedsymbol\Gbar{{\bar G}}
\accentedsymbol\Hbar{{\bar H}}
\accentedsymbol\Ibar{{\bar I}}
\accentedsymbol\Jbar{{\bar J}}
\accentedsymbol\Kbar{{\bar K}}
\accentedsymbol\Lbar{{\bar L}}
\accentedsymbol\Mbar{{\bar M}}
\accentedsymbol\Nbar{{\bar N}}
\accentedsymbol\Obar{{\bar O}}
\accentedsymbol\Pbar{{\bar P}}
\accentedsymbol\Qbar{{\bar Q}}
\accentedsymbol\Rbar{{\bar R}}
\accentedsymbol\Sbar{{\bar S}}
\accentedsymbol\Tbar{{\bar T}}
\accentedsymbol\Ubar{{\bar U}}
\accentedsymbol\Vbar{{\bar V}}
\accentedsymbol\Wbar{{\bar W}}
\accentedsymbol\Xbar{{\bar X}}
\accentedsymbol\Ybar{{\bar Y}}
\accentedsymbol\Zbar{{\bar Z}}

\accentedsymbol\abar{{\bar a}}
\accentedsymbol\bbar{{\bar b}}
\accentedsymbol\cbar{{\bar c}}
\accentedsymbol\dbar{{\bar d}}
\accentedsymbol\ebar{{\bar e}}
\accentedsymbol\fbar{{\bar f}}
\accentedsymbol\gbar{{\bar g}}
\makeatletter
\@ifundefined{hbar}{}{
        \let\hbar\@undefined
}
\makeatother
\accentedsymbol\hbar{{\bar h}}
\accentedsymbol\ibar{{\bar i}}
\accentedsymbol\jbar{{\bar j}}
\accentedsymbol\kbar{{\bar k}}
\accentedsymbol\lbar{{\bar l}}
\accentedsymbol\mbar{{\bar m}}
\accentedsymbol\nbar{{\bar n}}
\makeatletter
\@ifundefined{obar}{}{
        \let\obar\@undefined      
}
\makeatother
\accentedsymbol{\obar}{{\bar o}}        
\accentedsymbol\pbar{{\bar p}}
\accentedsymbol\qbar{{\bar q}}
\accentedsymbol\rbar{{\bar r}}
\accentedsymbol\sbar{{\bar s}}
\accentedsymbol\tbar{{\bar t}}
\accentedsymbol\ubar{{\bar u}}
\accentedsymbol\vbar{{\bar v}}
\accentedsymbol\wbar{{\bar w}}
\accentedsymbol\xbar{{\bar x}}
\accentedsymbol\ybar{{\bar y}}
\accentedsymbol\zbar{{\bar z}}

\accentedsymbol\mAhat{{\hat\mA}}
\accentedsymbol\mBhat{{\hat\mB}}
\accentedsymbol\mChat{{\hat\mC}}
\accentedsymbol\mDhat{{\hat\mD}}
\accentedsymbol\mEhat{{\hat\mE}}
\accentedsymbol\mFhat{{\hat\mF}}
\accentedsymbol\mGhat{{\hat\mG}}
\accentedsymbol\mHhat{{\hat\mH}}
\accentedsymbol\mIhat{{\hat\mI}}
\accentedsymbol\mJhat{{\hat\mJ}}
\accentedsymbol\mKhat{{\hat\mK}}
\accentedsymbol\mLhat{{\hat\mL}}
\accentedsymbol\mMhat{{\hat\mM}}
\accentedsymbol\mNhat{{\hat\mN}}
\accentedsymbol\mOhat{{\hat\mO}}
\accentedsymbol\mPhat{{\hat\mP}}
\accentedsymbol\mQhat{{\hat\mQ}}
\accentedsymbol\mRhat{{\hat\mR}}
\accentedsymbol\mShat{{\hat\mS}}
\accentedsymbol\mThat{{\hat\mT}}
\accentedsymbol\mUhat{{\hat\mU}}
\accentedsymbol\mVhat{{\hat\mV}}
\accentedsymbol\mWhat{{\hat\mW}}
\accentedsymbol\mXhat{{\hat\mX}}
\accentedsymbol\mYhat{{\hat\mY}}
\accentedsymbol\mZhat{{\hat\mZ}}

\accentedsymbol\vahat{{\hat\va}}
\accentedsymbol\vbhat{{\hat\vb}}
\accentedsymbol\vchat{{\hat\vc}}
\accentedsymbol\vdhat{{\hat\vd}}
\accentedsymbol\vehat{{\hat\ve}}
\accentedsymbol\vfhat{{\hat\vf}}
\accentedsymbol\vghat{{\hat\vg}}
\accentedsymbol\vhhat{{\hat\vh}}
\accentedsymbol\vihat{{\hat\vi}}
\accentedsymbol\vjhat{{\hat\vj}}
\accentedsymbol\vkhat{{\hat\vk}}
\accentedsymbol\vlhat{{\hat\vl}}
\accentedsymbol\vmhat{{\hat\vm}}
\accentedsymbol\vnhat{{\hat\vn}}
\accentedsymbol\vohat{{\hat\vo}}
\accentedsymbol\vphat{{\hat\vp}}
\accentedsymbol\vqhat{{\hat\vq}}
\accentedsymbol\vrhat{{\hat\vr}}
\accentedsymbol\vshat{{\hat\vs}}
\accentedsymbol\vthat{{\hat\vt}}
\accentedsymbol\vuhat{{\hat\vu}}
\accentedsymbol\vvhat{{\hat\vv}}
\accentedsymbol\vwhat{{\hat\vw}}
\accentedsymbol\vxhat{{\hat\vx}}
\accentedsymbol\vyhat{{\hat\vy}}
\accentedsymbol\vzhat{{\hat\vz}}

\newif\iflong
\longtrue

\definecolor{Worker1}{HTML}{1f78b4}
\definecolor{Worker1light}{HTML}{a6cee3}
\definecolor{Worker2}{HTML}{33a02c}
\definecolor{Worker2light}{HTML}{b2df8a}
\definecolor{Worker3}{HTML}{DCE319}
\definecolor{neutralbg}{HTML}{eeeeee}

\definecolor{Key3}{HTML}{481769}
\definecolor{Key4}{HTML}{D8E219}
\usepackage{pgfplots}
\usepackage{xstring} 
\usepgfplotslibrary{colormaps} 
\usetikzlibrary{pgfplots.colormaps} 
\usetikzlibrary{shapes.geometric}
\usetikzlibrary{decorations.text}
\pgfplotsset{colormap/viridis}
\usetikzlibrary[pgfplots.colormaps] 
\usetikzlibrary{shapes,snakes,shapes.geometric,calc,arrows,matrix,decorations.pathreplacing}

\newcommand*\circled[1]{\tikz[baseline=(char.base),every node/.style={scale=0.8}]{
    \node[shape=circle,draw,inner sep=1.0pt, fill=black, text=white] (char) {\textbf{#1}};}}

\newcommand*\circledC[2]{\tikz[baseline=(char.base),every node/.style={scale=0.8}]{
    \node[shape=circle,draw=#1,inner sep=1.0pt, fill=#1, text=white] (char) {\textbf{#2}};}}

\def \routfontsize {\normalsize}

\tikzset{
	square/.style={regular polygon,regular polygon sides=4,inner sep=0mm}, 
	object/.style={square,text width = 0.9cm,draw=black, align=center,fill=neutralbg, rounded corners,font=\routfontsize}, 
	object_outlier/.style={square,text width = 0.9cm,draw=white, align=center,fill=white, text=white,rounded corners,font=\routfontsize}, 
	arrow_style/.style={->,>=stealth',line width=0.2mm, shorten <=4pt, shorten >=4pt}, 
	arrow_caption/.style={color=black,align=center,font=\routfontsize },
	description/.style={align=left, font=\routfontsize}
}

\tikzset{	
	col_object/.style={minimum width=4.9cm, minimum height = 2.8cm,draw=black, rounded corners, fill=neutralbg, align=center}, 
	col_arrow_style/.style={<->,>=stealth',line width=0.2mm, color=black, shorten <=4pt, shorten >=4pt}, 
	col_arrow_caption/.style={sloped, color=black,align=center,font=\scriptsize }, 
	col_rectangles/.style={draw=black,fill=Worker2light, rounded corners=0.8mm,minimum height=4.5mm,minimum width=3.2cm]},
	col_small_arrows/.style={<->,>=stealth', line width=0.26mm, shorten <=2pt, shorten >=2pt}
}

\def\xlist{4}
\def\ylist{4}

\newcommand{\fillrandom}[5]{
	\pgfmathsetmacro{\w}{#2}
	\pgfmathsetmacro{\z}{#4}
	\foreach \y in {\w,...,\z}{
			\foreach \i in {0,1,2}{
        \pgfmathsetmacro{\x}{random(int(#1),int(#3))/10}
		        \xdef\collision{0}
		\ifnum\collision=0
            \xdef\xlist{\xlist,\x}
            \xdef\ylist{\ylist,\y}
		\path [fill=#5] (\x,\y/10) rectangle +(1/10,1/10);
        \fi 
	} }
}

\newcommand\sys{\textsc{Lapse}}

\newcommand\figurespace{\vspace{-0.29cm}}

\newcommand{\inputtikz}[1]{%
  \includegraphics[page=1]{figures/#1.pdf}%
}

\newtheorem{theorem}{Theorem}

\newcommand\rectM{$10\text{m}\times1\text{m}$}
\newcommand\sqM{$3.4\text{m} \times 3\text{m}$}

\begin{document}

\maketitle

\begin{abstract}
  To keep up with increasing dataset sizes and model complexity, distributed
  training has become a necessity for large machine learning tasks. Parameter
  servers ease the implementation of distributed parameter management---a key
  concern in distributed training---, but can induce severe communication
  overhead. To reduce communication overhead, distributed machine learning algorithms use
  techniques to increase parameter access locality (PAL), achieving up to linear
  speed-ups. We found that existing parameter servers provide only limited
  support for PAL techniques, however, and therefore prevent efficient training.
  In this paper, we explore whether and to what extent PAL techniques can be
  supported, and whether such support is beneficial. We propose to integrate
  dynamic parameter allocation into parameter servers, describe an efficient
  implementation of such a parameter server called \sys{}, and experimentally
  compare its performance to existing parameter servers across a number of
  machine learning tasks. We found that \sys{} provides near-linear scaling and
  can be orders of magnitude faster than existing parameter servers.
\end{abstract}

\section{Introduction}

\begin{figure}[t]
  \centering
  \includegraphics[page=1,width=1\columnwidth]{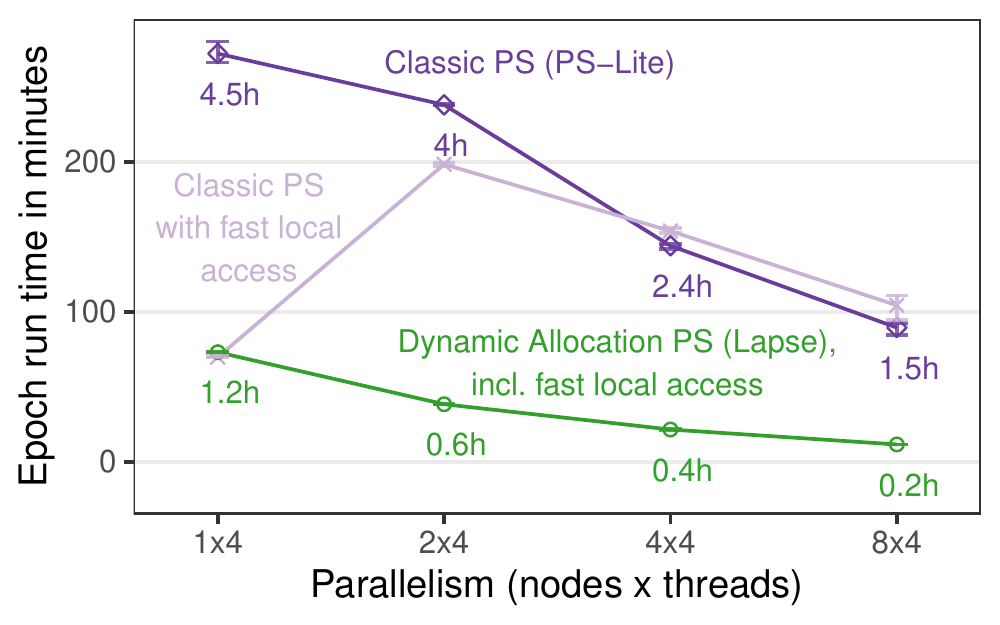}
  \caption{Parameter server (PS) performance for a large knowledge graph
    embeddings task (RESCAL, dimension~100). The performance of the classic PSs
    falls behind the performance of a single node due to communication overhead. In
    contrast, dynamic parameter allocation enables \sys{} to scale
    near-linearly. Details in Section~\ref{sec:experiments:setup}.}
  \label{fig:first-page}
  \figurespace{}
\end{figure}


To keep up with increasing dataset sizes and model complexity, distributed
training has become a necessity for large machine learning (ML) tasks.
Distributed ML allows (1) for models and data larger than the memory of a
  single machine, and (2) for faster training by leveraging distributed
  compute. In distributed ML, both training data and model parameters are
partitioned across a compute cluster. Each node in the cluster usually accesses
only its local part of the training data, but reads and/or updates most of the
model parameters. Parameter management is thus a key concern in distributed ML.
Applications either manage model parameters manually using low-level distributed
programming primitives or delegate parameter management to a \emph{parameter
  server} (PS). PSs provide primitives for reading and writing parameters and
handle partitioning and synchronization across nodes. Many ML stacks use PSs as
a component, e.g., TensorFlow~\cite{tensorflow}, MXNet~\cite{mxnet}, PyTorch
BigGraph~\cite{biggraph}, STRADS~\cite{strads}, STRADS-AP~\cite{stradsap}, or
Project Adam~\cite{adam}, and there exist multiple standalone PSs, e.g.,
Petuum~\cite{ssp}, PS-Lite~\cite{pslite}, Angel~\cite{angel},
FlexPS~\cite{flexps}, Glint~\cite{glint}, and PS2~\cite{ps2}.

As parameters are accessed by multiple nodes in the cluster and therefore
need to be transferred between nodes, distributed ML algorithms may suffer from
severe communication overhead when compared to single-machine implementations.
Figure~\ref{fig:first-page} shows exemplarily that the performance of a
distributed ML algorithm may fall behind the performance of single machine
algorithms when a classic PS such as PS-Lite is used. To reduce the impact of
communication, distributed ML algorithms employ techniques~\cite{dsgd, nomad,
  dsgdpp, flexifact, biggraph, nomad-mlr, nomad-lda, harp-lda+,
  graphlab-distributed, powergraph, cerebro} that increase \emph{parameter
  access locality} (PAL) and can achieve linear speed-ups. Intuitively, PAL
techniques ensure that most parameter accesses do not require (synchronous)
communication; example techniques include exploiting natural clustering of data,
parameter blocking, and latency hiding. Algorithms that use PAL techniques
typically manage parameters manually using low-level distributed programming
primitives.

Most existing PSs are easy to use---there is no need for low-level distributed
programming---, but provide only limited support for PAL techniques. One
limitation, for example, is that they allocate parameters statically. Moreover,
existing approaches to reducing communication overhead in PSs provide only
limited scalability compared to using PAL techniques (e.g., replication and
bounded staleness~\cite{ssp, essp}) or are not applicable to the ML algorithms
that we study (e.g., dynamically reducing cluster size~\cite{flexps}).

In this paper, we explore whether and to what extent PAL techniques can be
supported in PSs, and whether such support is beneficial. To improve PS
performance and suitability, we propose to integrate \emph{dynamic parameter
  allocation} (DPA) into PSs. DPA dynamically allocates parameters where they
are accessed, while providing location transparency and PS consistency
guarantees, i.e., sequential consistency. By doing so, PAL techniques can be
exploited directly. We discuss design options for PSs with DPA and describe an
efficient implementation of such a PS called \sys{}.

Figure~\ref{fig:first-page} shows the performance of \sys{} for the task of
training knowledge graph embeddings~\cite{rescal} using data clustering and
latency hiding PAL techniques. In contrast to classic PSs, \sys{} outperformed
the single-machine baseline and showed near-linear speed-ups. In our
experimental study, we observed similar results for multiple other ML tasks
(matrix factorization and word vectors): the classic PS approach barely
outperformed the single-machine baseline, whereas \sys{} scaled near-linearly,
with speed-ups of up to two orders of magnitude compared to classic PSs and up
to one order of magnitude compared to state-of-the-art~PSs.
Figure~\ref{fig:first-page} further shows that---although critical to the
  performance of \sys{}---fast local access alone does not alleviate the
  communication overhead of the classic PS approach.

In summary, our contributions are as follows. (i) We examine whether and to what
extent existing PSs support using PAL techniques to reduce communication
overhead. (ii) We propose to integrate DPA into PSs to be able to support PAL
techniques directly. (iii) We describe \sys{}, an efficient implementation of a
PS with DPA. (iv) We experimentally investigate the efficiency of classic PSs,
PSs with bounded staleness, and \sys{} on a number of ML tasks.

\begin{figure}[t]
    \centering
    \resizebox{1.0\columnwidth}{!}{
      \inputtikz{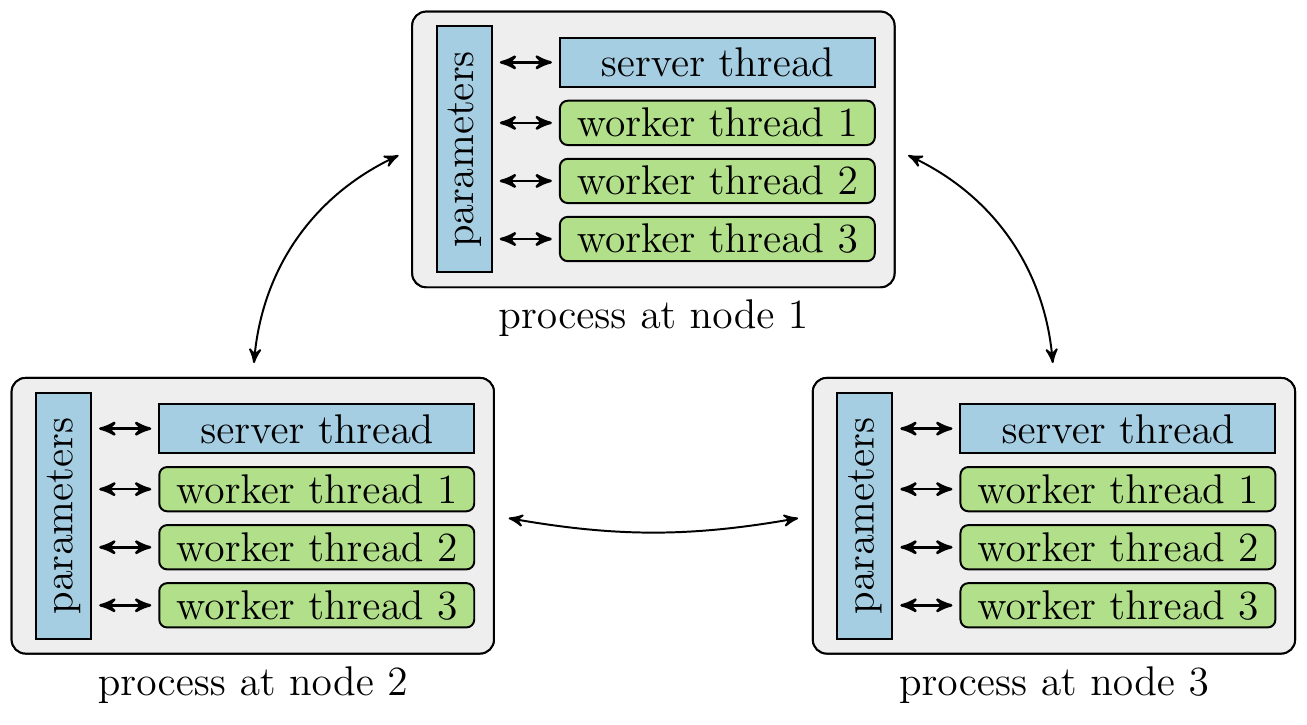}
    }
    \caption{PS architecture with server and worker threads
      co-located in one process per node. \sys{} employs this architecture.}
    \label{fig:colocation}
    \figurespace{}
\end{figure}


\section{The Case for Dynamic\\Parameter Allocation}
\label{sec:prelims}

We start by reviewing basic PS architectures (Section~\ref{sec:ps}). Second, we
outline common PAL techniques used in distributed ML (Section~\ref{sec:pal}).
For each technique, we discuss to what extent it is supported in existing PSs
and identify which features would be required to enable or improve support.
Finally, we introduce DPA, which enables PSs to exploit PAL techniques directly
(Section~\ref{sec:dpa}).

\subsection{Basic PS Architectures}
\label{sec:ps}

PSs~\cite{smola10, ahmed12, distbelief, ssp, pslite} partition the model
parameters across a set of \emph{servers}. The training data are usually
partitioned across a set of \emph{workers}. During training, each worker
processes its local part of the training data (often multiple times) and
continuously reads and updates model parameters. To coordinate parameter
accesses across workers, each parameter is assigned a unique \emph{key} and the
PS provides \texttt{pull} and \texttt{push} primitives for reads and writes,
respectively; cf.~Table~\ref{tab:api}. Both operations can be performed
synchronously or asynchronously. The push operation is usually cumulative, i.e.,
the client sends an update term to the PS, which then adds this term to the
parameter value.

Although servers and workers may reside on different machines, they are often
co-located for efficiency reasons (especially when PAL techniques are used).
Some architectures~\cite{pslite, glint, angel} run one server process and one or
more worker processes on each machine, others~\cite{ssp, flexps} use a single
process with one server thread and multiple worker threads to reduce
inter-process communication. Figure~\ref{fig:colocation} depicts such a PS
architecture with one server and three worker threads per node.

\newcommand{\yes}{\checkmark}
\newcommand{\no}{\boldmath$\times$}
\newcommand{\lcnote}{}
\newcommand{\nwnote}{\tnote{{b}}}
\newcommand{\nsp}[1]{\hspace{-0.1cm}#1\hspace{-0.1cm}}

\begin{table}
  \caption{Per-key consistency guarantees of PS architectures, using
    representatives for types: PS-Lite~\protect\cite{pslite} for classic and
    Petuum~\protect\cite{petuum} for stale.}
  \label{tab:consistency}
  \scriptsize
  \centering
  \begin{threeparttable}
    \begin{tabular}{lcccccc}
      \toprule

      Parameter Server & \multicolumn{2}{c}{Classic} & \multicolumn{3}{c}{\sys{}} &  Stale \\
      \cmidrule(lr){2-3} \cmidrule(lr){4-6} \cmidrule(lr){7-7}
      Synchronous          & \nsp{sync} & \nsp{async} & sync & \multicolumn{2}{c}{async}  & sync, async \\
       \cmidrule(lr){5-6}
        Location caches        &  &                &       & \nsp{off}  & \nsp{on} &              \\
      \midrule
      Eventual & \yes & \yes & \yes & \yes & \yes & \yes \\
      PRAM\tnote{a}~\cite{pramconsistency} & \yes & \yes\nwnote & \yes & \yes\nwnote & \no \lcnote & \yes \\
      Causal~\cite{causalconsistency} & \yes & \yes\nwnote & \yes & \yes\nwnote & \no \lcnote & \no  \\
      Sequential~\cite{sequentialconsistency} & \yes & \yes\nwnote & \yes & \yes\nwnote & \no \lcnote & \no  \\
      Serializability & \no & \no & \no & \no & \no & \no  \\

      \bottomrule
    \end{tabular}
    \begin{tablenotes}
    \item[a] I.e., monotonic reads, monotonic writes, and read your writes
    \item[b] Assuming that the network layer preserves message order (which is
      true for \sys{} and PS-Lite)
    \end{tablenotes}
  \end{threeparttable}
  \figurespace{}
\end{table}


\begin{figure*}[t]
  \centering
  \resizebox{0.8\textwidth}{!}{
    \inputtikz{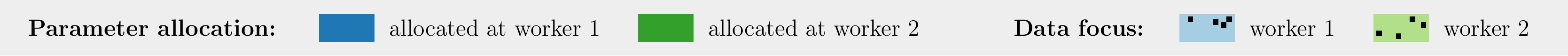}
    }
   \hspace{0.2cm}
   \vspace{0.3cm}

  \begin{subfigure}[b]{.23\textwidth}
    \centering
    \resizebox{1.0\columnwidth}{!}{
    \inputtikz{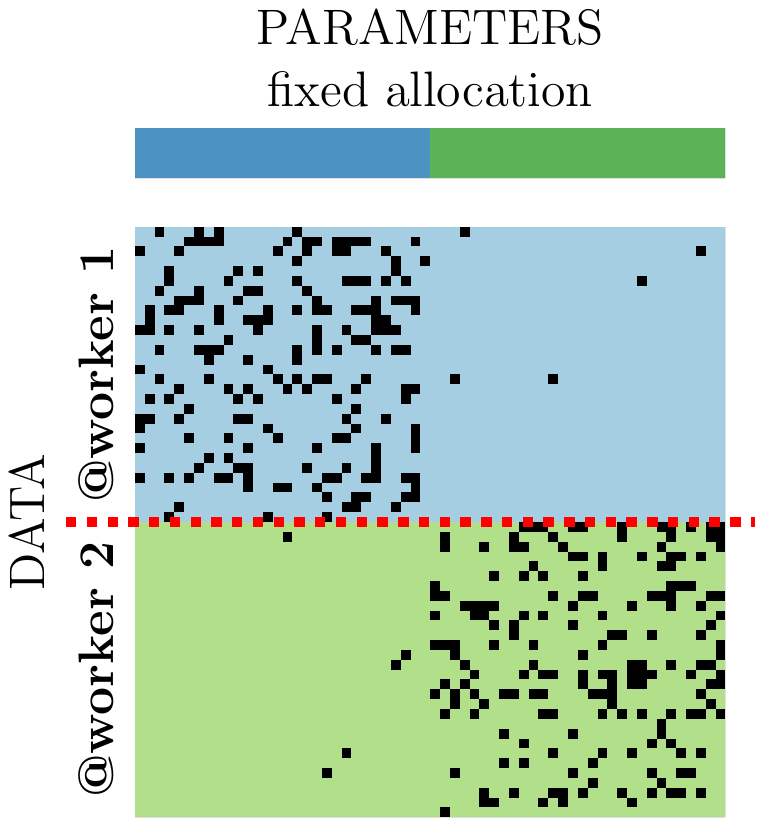}
	}
  \caption{Data clustering}
    \label{fig:locality:data}
  \end{subfigure}%
  \begin{subfigure}[b]{.45\textwidth}
    \centering
    \resizebox{1.0\columnwidth}{!}{
    \inputtikz{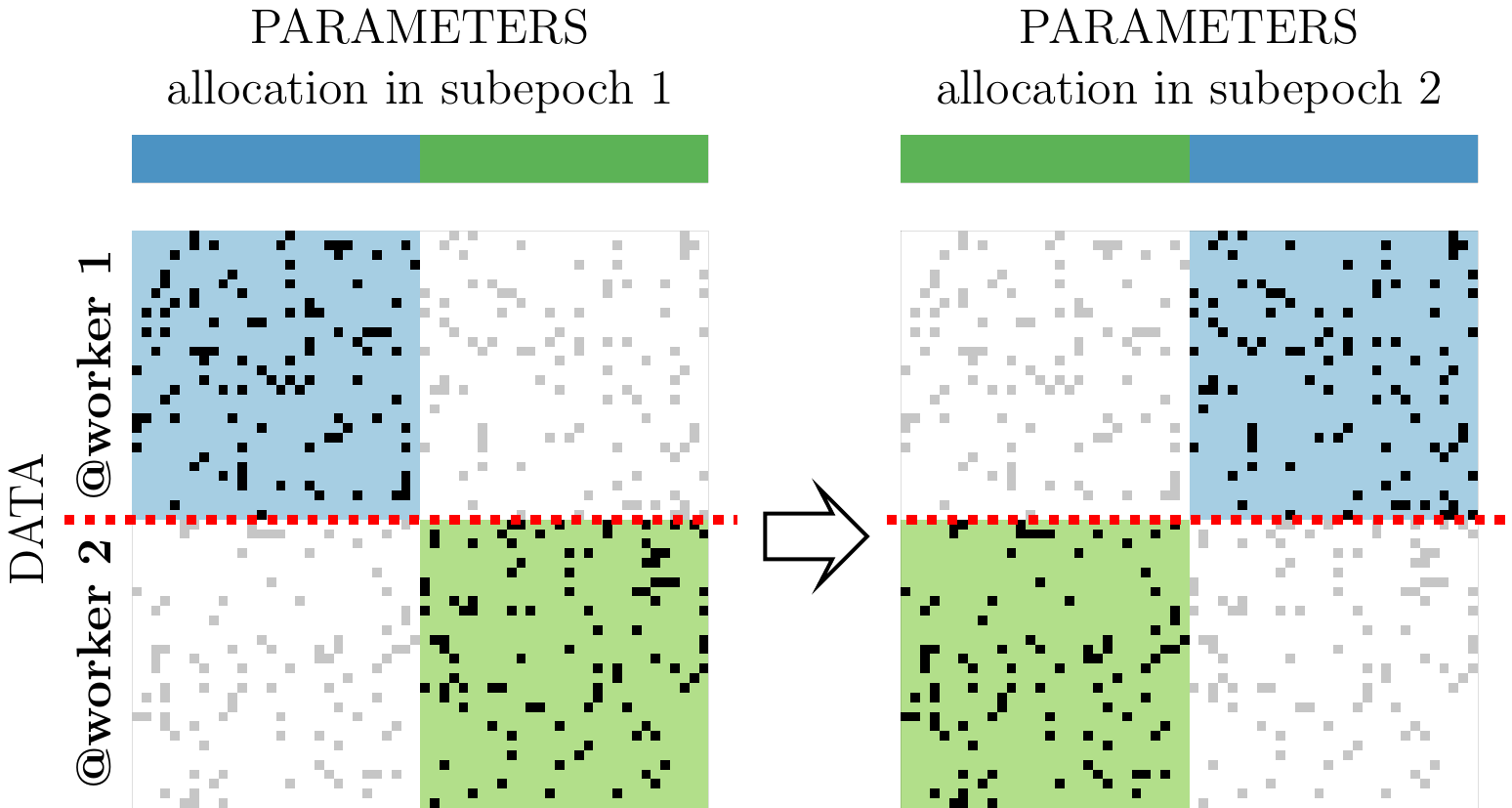}
}
\caption{Parameter blocking}
    \label{fig:locality:ca}
  \end{subfigure}%
  \begin{subfigure}[b]{.27\textwidth}
    \centering
    \resizebox{1.0\columnwidth}{!}{
    \inputtikz{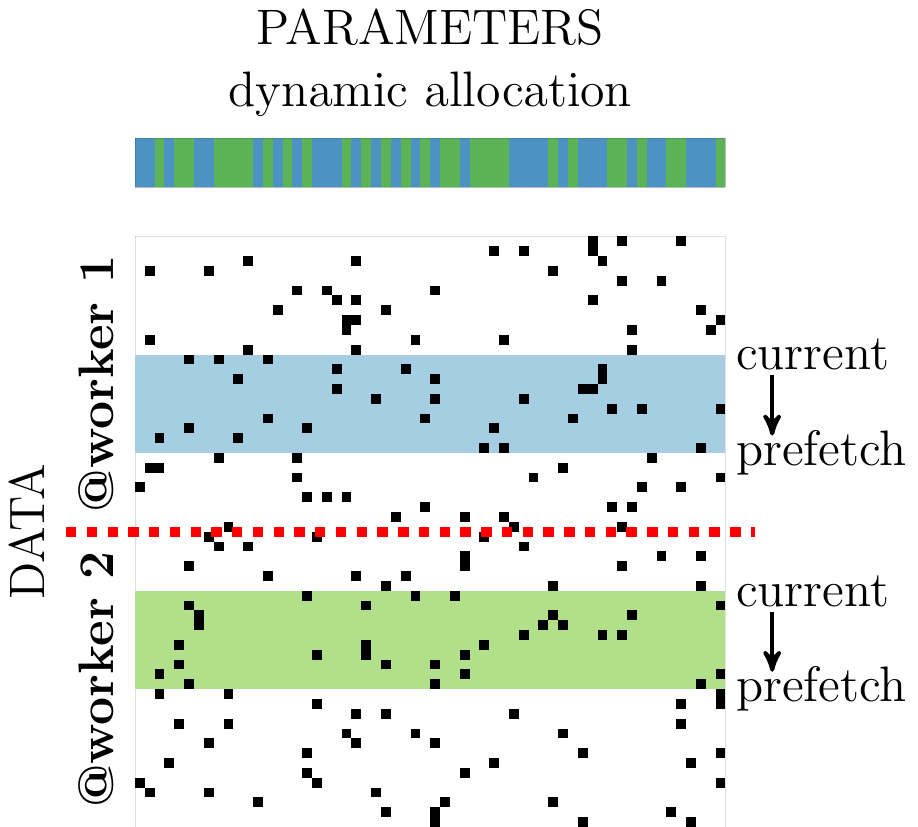}
}
\caption{Latency hiding}
    \label{fig:locality:lh}
  \end{subfigure}
  \caption{Techniques to reduce communication cost, based on parameter access
    locality: workers access different subsets of the model parameters over
    time. Rows correspond to data points, columns to parameters, black dots to
    parameter accesses. (a) Training data are clustered such that each worker
    accesses mostly a separate subset of parameters
    (Section~\ref{sec:data-clustering}). (b) Within each subepoch, each worker
    is restricted to one block of parameters. Which worker has access to which
    block changes from subepoch to subepoch (Section~\ref{sec:blocking}). (c)
    Asynchronously prefetching (or prelocalizing) of parameter values, such that
    they can be accessed locally, hides access latency
    (Section~\ref{sec:latency-hiding}).}
  \figurespace{}
  \label{fig:locality}
\end{figure*}


In the \emph{classic PS} architecture, parameters are statically allocated to
servers (e.g., via a range partitioning of the parameter keys) and there is no
replication. Thus precisely one server holds the current value of a parameter,
and this server is used for all pull and push operations on this parameter.
Classic PSs typically guarantee sequential
consistency~\cite{sequentialconsistency} for operations on the same key. This
means that (1) each worker's operations are executed in the order specified by
the worker, and (2) the result of any execution is equivalent to an execution of
the operations of all workers in some sequential order. Note that lost updates
do not occur in PSs when updates are cumulative. Table~\ref{tab:consistency}
gives an overview of provided consistency guarantees for different types of PSs.
PSs give no guarantees across multiple keys.

The \emph{stale PS} architecture employs replication and
tolerates some amount of staleness in the replicas~\cite{ssp, flexps, angel,
  essp, iterstore}. In such architectures, parameters are still statically allocated to
servers as in a classic PS, but the PS may replicate a subset of the parameters
to additional servers to reduce communication overhead~\cite{ssp}. This is
beneficial especially when servers and workers are co-located because parameters
can be replicated to the subset of servers that access them. Stale PSs often
provide weaker forms of consistency than classic PSs (PRAM consistency
or only eventual consistency). They commonly do support \emph{bounded staleness}
and require applications to explicitly control staleness via special primitives
(e.g., an ``advance the clock'' operation).

\subsection{PAL Techniques}
\label{sec:pal}

We consider three common PAL techniques: data clustering, parameter blocking,
and latency hiding.

\subsubsection{Data Clustering}
\label{sec:data-clustering}

One method to reduce communication cost is to exploit structure in training
data~\cite{graphlab-distributed, powergraph, powerlyra, smola10, ahmed12}. For
example, consider a training data set that consists of documents written in two
different languages and an ML model that associates a parameter with each word
(e.g., a bag-of-words classifier or word vectors). When processing a document
during training, only the parameters for the words contained in the document are
relevant. This property can be exploited using \emph{data clustering}. For
example, if a separate worker is used for the documents of each language,
different workers access mostly separate parameters. This is an example of PAL:
different workers access different subsets of the parameters at a given time.
This locality can be exploited by allocating parameters to the worker machines
that access them. Figure~\ref{fig:locality:data} depicts an example; here rows
correspond to documents, dots to words, and each parameter is allocated to the
node where it is accessed most frequently.

Data clustering can be exploited in existing PSs in principle, although it is
often painful to do so because PSs provide no direct control over the allocation
of the parameters. Instead, parameters are typically partitioned using either
hash or range partitioning. To exploit data clustering, applications may
manually enforce the desired allocation by \emph{key design}, i.e., by
explicitly assigning keys to parameters such that the parameters are allocated
to the desired node. Such an approach requires knowledge of PS internals,
preprocessing of the training data, and a custom implementation for each task.
To improve support for data clustering, PSs should provide support for explicit
\textbf{parameter \mbox{location control}}.

To exploit data clustering, it is essential that the PS provides \textbf{fast
  access to local parameters}; e.g., by using shared memory as in manual
implementations~\cite{dsgd, nomad}. However, to the best of our knowledge, all
existing PSs access parameters either through inter-process~\cite{pslite} or
inter-thread communication~\cite{petuum}, leading to overly high access latency.

\subsubsection{Parameter Blocking}
\label{sec:blocking}

An alternative approach to provide PAL is to divide the model parameters into
blocks. Training is split into \emph{subepochs} such that each worker is
restricted to one block of parameters within each subepoch. Which worker has
access to which block changes from subepoch to subepoch, however. Such
\emph{parameter blocking} approaches have been developed for many ML algorithms,
including  matrix factorization~\cite{dsgd, nomad, dsgdpp}, tensor
factorization~\cite{flexifact}, latent dirichlet allocation~\cite{nomad-lda,
  harp-lda+}, multinomial logistic regression~\cite{nomad-mlr}, and knowledge
graph embeddings~\cite{biggraph}. They were also proposed for efficient
multi-model training~\cite{cerebro}.

Manual implementations exploit parameter blocking by allocating parameters to
the node where they are currently accessed~\cite{dsgd,dsgdpp,nomad}, thus
eliminating network communication for individual parameter accesses.
Communication is required only between subepochs. Figure~\ref{fig:locality:ca}
depicts a simplified example for a matrix factorization task~\cite{dsgd}. In the
first subepoch, worker 1 (worker 2) focuses on the left (right) block of the
model parameters. It does so by only processing the corresponding part of its
data and ignoring the remainder. In the second subepoch, each worker processes
the other block and the other part of its data. This process is repeated multiple
times.

Existing PSs offer limited support for parameter blocking because they allocate
parameters \emph{statically}. This means that a parameter is assigned to one
server and stays there throughout training. It is therefore not possible to
dynamically allocate a parameter to the node where it is currently accessed.
Parameter blocking can be emulated to some extent in stale PS architectures,
however. This requires the creation of replicas for each block and forced
refreshes of replicas between subepochs. Such an approach is limited to
synchronous parameter blocking approaches, requires changes to the
implementation, and induces unnecessary communication (because parameters are
transferred via their server instead of directly from worker to worker). To
exploit parameter blocking efficiently, PSs need to support \textbf{parameter
  relocation}, i.e., the ability to move parameters among nodes during run time.

\subsubsection{Latency Hiding}
\label{sec:latency-hiding}

Latency hiding techniques reduce communication overhead (but not communication
itself). For example, prefetching is commonly used when there is a distinction
between local and remote data, such as in processor caches~\cite{cachememories}
or distributed systems~\cite{steen17}. In distributed ML, the latency of
parameter access can be reduced by ensuring that a parameter value is already
present at a worker at the time it is accessed~\cite{essp,iterstore,dsgdpp}. Such an
approach is beneficial when parameter access is sparse, i.e., each worker
accesses few parameters at a time.

Prefetching can be implemented by pulling a parameter \emph{asynchronously}
before it is needed. The disadvantage of this approach is that an application
needs to manage prefetched parameters, and that updates that occur between
prefetching a parameter and using it are not visible. Therefore, such an
approach provides neither sequential nor causal consistency
(cf.~Table~\ref{tab:consistency}). Moreover, the exchange of parameters between
different workers always involves the server, which may be inefficient. An
alternative approach is the ESSP consistency protocol of Petuum~\cite{essp},
which proactively replicates all previously accessed parameters at a node
(during its ``clock'' operation). This approach avoids parameter management, but
does not provide sequential consistency.

An alternative to prefetching is to \emph{prelocalize} a parameter before
access, i.e., to reallocate the parameter from its current node to the node
where it is accessed and to \emph{keep it there} afterward (until it is
prelocalized by some other worker). This approach is illustrated in
Figure~\ref{fig:locality:lh}. Note that, in contrast to prefetching, the parameter
is not replicated. Consequently, parameter updates by other workers are
immediately visible after prelocalization. Moreover, there is no need to write
local updates back to a remote location as the parameter is now stored
locally. To support prelocalization, PSs need to support \textbf{parameter
  relocation with consistent access} before, during, and after relocation.

\subsection{Dynamic Parameter Allocation}
\label{sec:dpa}

As discussed above, existing PSs offer limited support for the PAL techniques of
distributed ML. The main obstacles are that existing PSs provide limited control
over parameter allocation and perform allocation statically. In more detail, we
identified the following requirements to enable or \mbox{improve support}:
\begin{description}
\item[Fast local access.] PSs should provide low-latency access to local
  parameters.
  \vspace{-0.2cm}
\item[Parameter location control.] PSs should allow applications to control where
  a parameter is stored.
  \vspace{-0.2cm}
\item[Parameter relocation.] PSs should support relocating parameters between
  servers during runtime.
  \vspace{-0.2cm}
\item[Consistent access.] Parameter access should be consistent before, during,
  and after a relocation.
\end{description}
To satisfy these requirements, the PS must support DPA, i.e., it must be able to
change the allocation of parameters during runtime. While doing so, the PS
semantics must not change: \texttt{pull} and \texttt{push} operations need to be
oblivious of a parameter's current location and provide correct results whether
or not the parameter is currently being relocated. This requires the PS to
manage parameter locations, to transparently route parameter accesses to the
parameter's current location, to handle reads and writes correctly during
relocations, and to provide to applications new primitives to initiate
relocations.

A DPA PS enables support for PAL techniques roughly as follows: each worker
instructs the PS to \emph{localize} the parameters that it will access
frequently in the near future, but otherwise uses the PS as it would use any
other PS, i.e., via the \texttt{pull} and \texttt{push} primitives. For data
clustering, applications control parameter locations once in the beginning: each
node localizes the parameters that it accesses more frequently than the other
nodes. Subsequently, the majority of parameter accesses (using \texttt{pull} and
\texttt{push}) is local. For parameter blocking, at the beginning of each
subepoch, applications move parameters of a block to the node that accesses them
during the subepoch. Parameter accesses (both reads and writes) within the
subepoch then require no further network communication. Finally, for latency
hiding, workers prelocalize parameters before accessing them. When the parameter
is accessed, latency is low because the parameter is already local (unless
another worker localized the parameter in the meantime). Concurrent updates by
other workers are seen locally, because the PS routes them to the parameter's
current location.


\begin{table*}
  \caption{Primitives of \sys{}, a PS with dynamic parameter allocation. The
    push primitive is cumulative. All primitives can run synchronously or
    asynchronously. Compared to classic PSs, \sys{} adds one primitive to
    initiate parameter relocations.}
  \label{tab:api}
  \centering
  \begin{threeparttable}
    \begin{tabular}{lccl}
      \toprule
      Primitive & \multicolumn{2}{c}{Support for} & Description \\
      \cmidrule(lr){2-3}
                & sync. & async. & \\
      \midrule
      \texttt{pull(parameters)}  & \checkmark & \checkmark & Retrieve the values of \texttt{parameters} from the corresponding servers. \\
      \texttt{push(parameters, updates)} & \checkmark & \checkmark & Send \texttt{updates} for \texttt{parameters} to the corresponding servers. \\
      \texttt{localize(parameters)}  & \checkmark & \checkmark & Request local allocation of  \texttt{parameters}.  \\
      \bottomrule
    \end{tabular}
  \end{threeparttable}
  \figurespace{}
\end{table*}


\section{The Lapse Parameter Server}
\label{sec:architecture}

To explore the suitability of PSs with DPA as well as architectural design
choices, we created \sys{}. \sys{} is based on PS-Lite and aims to fulfill the
requirements established in the previous sections. In particular, \sys{}
provides fast access to local parameters, consistency guarantees similar to
classic PSs, and efficient parameter relocation. We start with a brief overview
of \sys{} and subsequently discuss individual components, including parameter
relocation, parameter access, consistency, location management, granularity, and
important implementation aspects.

\subsection{Overview}
\label{sec:overview}

\sys{} co-locates worker and server threads, as illustrated in
Figure~\ref{fig:colocation}, because this
architecture facilitates low-latency local parameter access (see below).

\textbf{API.} \sys{} adds a single primitive called \texttt{localize} to the API
of the PS; see Table~\ref{tab:api}. The primitive takes the keys of one or more
parameters as arguments. When a worker issues a \texttt{localize}, it requests
that all provided parameters are relocated to its node. \sys{} then
transparently relocates these parameters and future accesses by the worker
require no further network communication. We opted for the \texttt{localize}
primitive---instead of a more general primitive that allows for relocation among
arbitrary nodes---because it is simpler and sufficiently expressive to support
PAL techniques. Furthermore, \texttt{localize} preserves the PS property that two
workers logically interact only via the servers (and not
directly)~\cite{pslite}. Workers access localized parameters in the same way as
non-localized parameters. This allows \sys{} to relocate parameters without
affecting workers that~use~them.

\textbf{Location management.} \sys{} manages parameter locations with a
decentralized home node approach: for each parameter, there is one \emph{owner
  node} that stores the current parameter value and one \emph{home node} that
knows the parameter's current location. The home node is assigned statically as
in existing PSs, whereas the owner node changes dynamically during run time. We
further discuss location management in Section~\ref{sec:management}.

\textbf{Parameter access.} \sys{} ensures that local parameter access is fast by
accessing local parameters via shared memory. For non-local parameter access,
\sys{} sends a message to the home node, which then forwards the message to the
current owner of a parameter. \sys{} optionally supports location caches, which
eliminate the message to the home node if a parameter is accessed repeatedly
while it is not relocated. See Section~\ref{sec:access} for details.

\textbf{Parameter relocation.} A localize call requires \sys{} to relocate the
parameter to the new owner and update the location information on the home node.
Care needs to be taken that push and pull operations that are issued while the
parameter is relocated are handled correctly. \sys{} ensures correctness by
forwarding all operations to the new owner immediately, possibly before the
relocation is finished. The new owner simply queues all operations until the
relocation is finished. \sys{} sends at most three messages for a relocation of
one parameter and pauses processing for the relocated parameter only for the
time that it takes to send one network message. The entire protocol is
\mbox{described in Section~\ref{sec:movement}}.

\textbf{Consistency.} In general, \sys{} provides the sequential consistency
guarantees of classic PSs even in the presence of parameter relocations. We show
in Section~\ref{sec:consistency} that the use of location caches may impact consistency
guarantees. In particular, when location caches are used, \sys{} still
provides sequential consistency for synchronous operations, but only eventual
consistency for asynchronous operations.

\subsection{Parameter Relocation}
\label{sec:movement}

A key component of \sys{} is the relocation of parameters. It is important that
this relocation is efficient because PAL techniques may relocate parameters
frequently (up to \num{36000} keys and \num{289} million
parameter values per second in our experiments).
We discuss how \sys{} relocates parameters, how it manages operations that are
issued during a relocation, and how it handles simultaneous relocation requests
by multiple nodes.

During a localize operation, (1) the home node needs to be informed of
the location change, (2) the parameter needs to be moved from its current owner
to the new owner, and (3) \sys{} needs to stop processing operations at the
current owner and start processing operations at the new owner. Key decisions
are what messages to send and how to handle operations that are issued during
parameter relocation. \sys{} aims to keep both the relocation time and the
blocking time for a relocation short. We use \emph{relocation time} to refer
to the time between issuing a localize call and the moment when the new
owner starts answering operations locally. By \emph{blocking time}, we mean the
time in which \sys{} cannot immediately answer operations for the parameter (but
instead queues operations for later processing). The two measures usually differ
because the current owner of the parameter continues to process operations for
some time after the localize call is issued at the requesting node,
i.e., the relocation time may be larger than the blocking time.

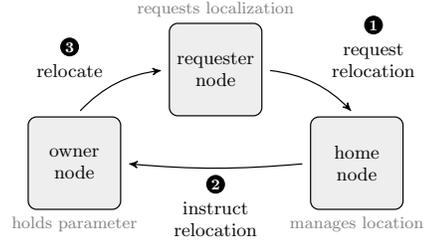
\begin{figure}
  \centering
  \resizebox{0.7\columnwidth}{!}{
    \begin{tikzpicture}[line width = 0.2mm,
	objekt/.style={square,minimum size=2.2cm,draw=black, align=center,fill=neutralbg, rounded corners,inner sep=-2mm, outer sep=0}, 
	square/.style={regular polygon,regular polygon sides=4,inner sep=0, outer sep=0}, 
	arrow_style/.style={->,>=stealth',line width=0.2mm, color=black, shorten <=4pt, shorten >=4pt}, 
	arrow_caption/.style={color=black,align=center,font=\normalsize },
	node description/.style={font=\small, color=gray},
	hidden shape/.style={white}
	]

	\matrix[row sep=0mm,column sep=8mm] {
		& \node [objekt](req) {requester\\node}; & \\
		\node [objekt](own) {owner\\node}; & & \node [objekt](home) {home\\node};
	\\};

	\draw [arrow_style](req.east) to [bend left=20] node[arrow_caption,above,xshift=1.0cm,yshift=-0.0cm] {\circled{1}\\[0.5mm] request\\ relocation} (home.north);
	\draw [arrow_style](home.west) to [bend left=5] node[arrow_caption,below] {\circled{2} \\[0.5mm] instruct \\relocation} (own.east);
	\draw [arrow_style](own.north) to [bend left=20] node[arrow_caption,above,xshift=-0.8cm,yshift=-0.0cm] {\circled{3} \\[0.5mm] relocate}(req.west);
	
	\draw (own.south) node[below, node description] {holds parameter};
	\draw (req.north) node[above, node description] {requests localization};
	\draw (home.south) node[below, node description] {manages location};


\end{tikzpicture}

  }
  \caption{A worker requests to localize a parameter. \sys{} transparently
    relocates the parameter from the current owner to the requester node and
    informs the home node of the location change.}
  \label{fig:localize-messages}
  \figurespace{}
\end{figure}

We refer to the node that issued the localize operation as the
\emph{requester node}. The requester node is the new owner of the parameter
after the relocation has finished. \sys{} sends three messages in total to relocate
a parameter, see Figure~\ref{fig:localize-messages}. \circled{1} The requester
node informs the home node of the parameter about the location change. The home
node updates the location information immediately and starts routing parameter
accesses for the relocated parameter to the requester node. \circled{2} The home
node instructs the old owner to stop processing parameter accesses for the relocated
parameter, to remove the parameter from its local storage, and to transfer it to the
requester node. \circled{3} The old owner hands over the parameter to the
requester node. The requester node inserts the parameter into its local storage
and starts processing parameter accesses for the relocated parameter.

\begin{figure*}[t]
  \begin{subfigure}[c]{.25\textwidth}
    \centering
    \resizebox{1.0\columnwidth}{!}{
    \inputtikz{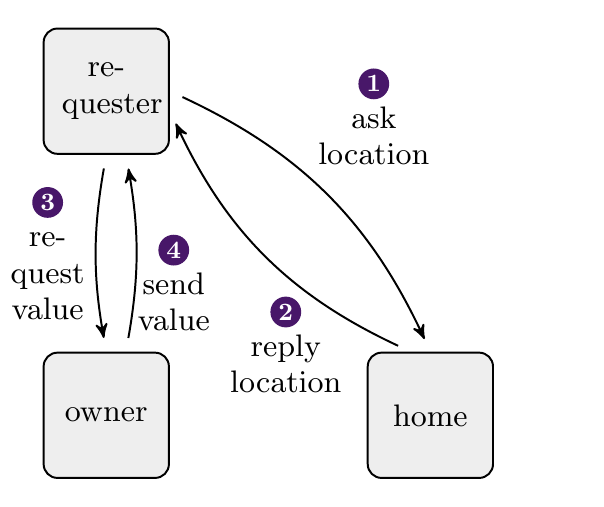}
	}
    \caption{Location request}
    \label{fig:routing:ask}
  \end{subfigure}%
  \begin{subfigure}[c]{.25\textwidth}
    \centering
    \resizebox{1.0\columnwidth}{!}{
    \inputtikz{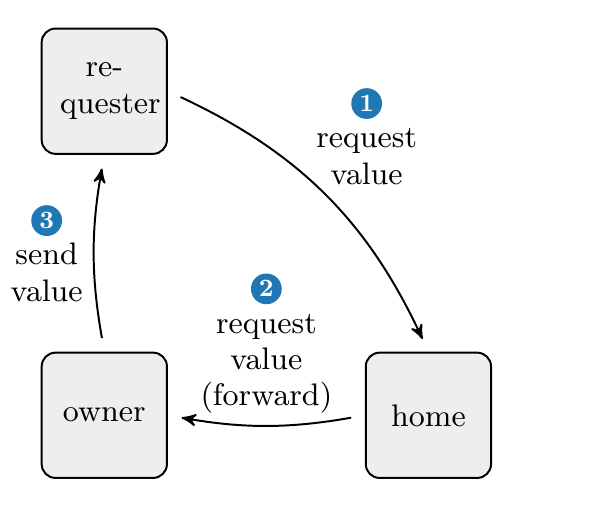}
	}
    \caption{Forward}
    \label{fig:routing:forward}
  \end{subfigure}%
  \begin{subfigure}[c]{.25\textwidth}
    \centering
    \resizebox{1.0\columnwidth}{!}{
    \inputtikz{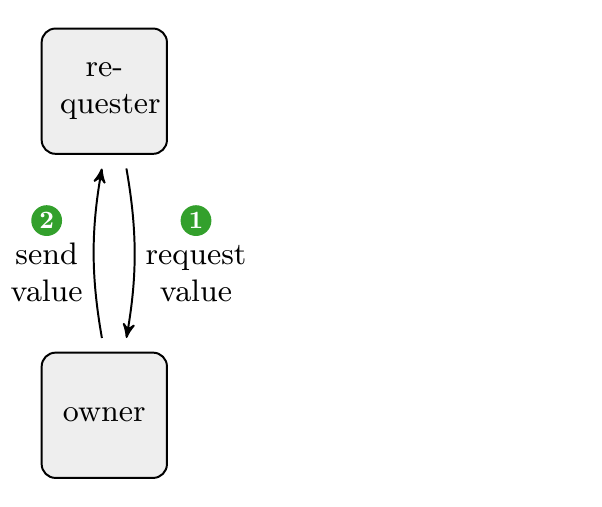}
	}
    \caption{Correct location cache}
    \label{fig:routing:hit}
  \end{subfigure}%
  \begin{subfigure}[c]{.25\textwidth}
    \centering
    \resizebox{1.0\columnwidth}{!}{    
\inputtikz{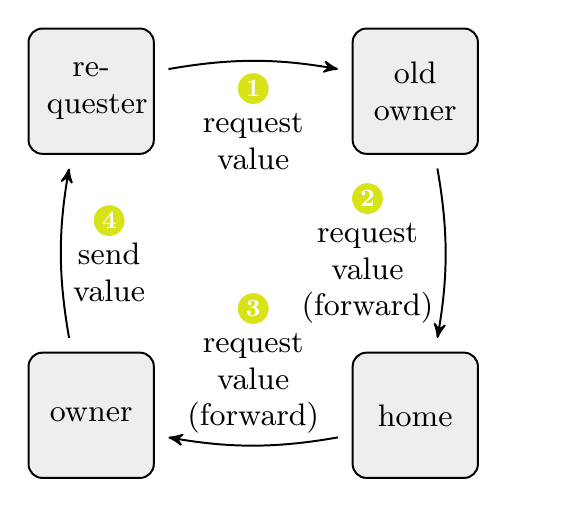}
}
    \caption{Stale cache: double-forward}
    \label{fig:routing:miss}
  \end{subfigure}%
  \caption{Routing for non-local parameter access. If the location of a
    parameter is unknown, \sys{} employs the \emph{forward} strategy (Figure~b),
    requiring 3 messages. \sys{} optionally supports location caches. A correct
    cache reduces the number of messages to 2 (Figure~c), a stale cache
    increases it to 4 (Figure~d). The figures depict labels for \texttt{pull}
    messages. Labels for \texttt{push} are analogously: \emph{request\,value}
    $\to$ \emph{send\,update}, \emph{send\,value} $\to$ \emph{confirm\,update},
    \emph{request\,value\,(forward)} $\to$ \emph{send\,update\,(forward)} }
  \label{fig:routing:overview}
  \figurespace{}
\end{figure*}


During the relocation, the requester node queues all parameter accesses that
involve the relocated parameter. It queues both local accesses (i.e., accesses by
workers at the requester node) and remote accesses that are routed to it before
the relocation is finished. Once the relocation is completed, it processes the
queued operations in order and then starts handling further accesses as the
new owner. As discussed in Section~\ref{sec:consistency}, this approach ensures that
sequential consistency is maintained.

In the absence of other operations, the relocation time for this protocol is
approximately the time for sending three messages over the network, and the
blocking time is the time for sending one message (because operations are queued
at the requester and the home node starts forwarding to the requester
immediately). One may try to reduce blocking time by letting the old owner
process operations until the relocation is complete (and forwarding all updates
to the new owner). However, such an approach would require additional network
communication and would increase relocation time. The protocol used by
\sys{} strikes a balance between short relocation and short blocking time.

If multiple nodes simultaneously localize the same parameter, there is a
\emph{localization conflict}: without replicas, a parameter resides at only one
node at a time. In the case of a localization conflict, the above protocol
transfers the parameter to each requesting node once (in the order the
relocation requests arrive at the home node). This gives each node a short
opportunity to process the parameter locally, but also causes communication
overhead for frequently localized parameters (because it repeatedly transfers
the parameter value, potentially in cycles). A short localize moratorium, in
which further localize requests are ignored, may reduce this cost, but would
change the semantics of the localize primitive, increase complexity, and may
impact overall efficiency. We did not consider such an approach in~\sys{}.

\subsection{Parameter Access}
\label{sec:access}

When effective PAL techniques are used, the majority of parameter accesses are
processed locally. Nevertheless, remote access to all parameters may arise at
all times and needs to be handled appropriately. We now discuss how \sys{}
handles local access, remote access, location caches, and access to a
parameter that is currently relocating.

\textbf{Local access.} \sys{} provides fast local parameter access by
accessing locally stored parameters via shared memory directly from the worker
threads, i.e., without involving the PS thread (see Figure~\ref{fig:colocation})
or other nodes. In
our experiments, accessing the parameter storage via shared memory provided up to 6x
lower latency than access via a PS thread using queues (as implemented in
Petuum~\cite{petuum}, for example).
As other PSs, \sys{} guarantees per-key atomic reads and writes; it
does so using latches (i.e., locks held for the time of the operation) for local \mbox{accesses (see Section~\ref{sec:implementation})}.

\textbf{Remote access.} We now discuss remote parameter access and first
assume that there is no location caching. There are two basic strategies. In the
\emph{location request} strategy, the worker retrieves the current owner of the
parameter from the home node and subsequently sends the pull or push request to
that owner (Figure~\ref{fig:routing:ask}). In the \emph{forward} strategy, the
worker sends the request itself to the home node, which then forwards it to the
current owner (Figure~\ref{fig:routing:forward}).
\sys{} employs the \emph{forward} strategy because (i) it always uses up-to-date
location information for routing decisions and (ii) it requires one message less
than \emph{location request}. The \emph{forward} strategy uses the latest location
information for routing because the home node, which holds the location
information, sends the request to the owner (message~\circledC{Worker1}{2}). In
contrast, in \emph{location request}, the requester node sends the request
(message~\circledC{Key3}{3}) based on the location obtained from the home node.
This location may be outdated if another worker requests a relocation after the
home node replied (message~\circledC{Key3}{2}). In this case, the requester node
may send message~\circledC{Key3}{3} to an outdated owner; such a case would
require special handling.

\textbf{Location caching.} \sys{} provides the option to cache the locations of
recently accessed parameters. This allows workers to contact the current owner
directly (Figure~\ref{fig:routing:hit}), reducing the number of necessary
messages to two. To avoid managing cached locations and sending invalidation
messages, the location caches are updated only after \texttt{push} and
\texttt{pull} operations and after parameter relocations (i.e., without any
additional messages). As a consequence, the cache may hold stale entries. If
such an entry is used, \sys{} uses a \emph{double-forward} approach, which
increases the number of sent messages by one
(Figure~\ref{fig:routing:miss}).

\textbf{Access during relocation.} Workers can issue operations for any
parameter at any time, including when a parameter is relocating. In the following,
we discuss how \sys{} handles different possible scenarios of operations on a
relocating parameter. First, suppose that the requester node (to which the parameter
is currently relocating) accesses the parameter. \sys{} then locally queues the
request at the requester node and processes it when the relocation is finished.
Second, suppose that the old owner accesses the parameter. \sys{} processes the
parameter access locally if it occurs before the parameter leaves the local
store. Otherwise, \sys{} sends the operation to the new owner and processes it
there. Finally, consider that a third node (neither the requester nor the old
owner) accesses the parameter. If location caches are disabled, there
are two cases. (1) The access arrives at the home node before the relocation. Then
\sys{} forwards the access to the old owner and processes it there (before the
relocation). (2) The access arrives at the home node after the relocation. Then \sys{}
forwards and processes it at the new owner. If necessary, the new owner queues
the access until the relocation is finished. With location caches, \sys{} additionally
processes the request at the old owner if the parameter's location is cached
correctly at the requester node and the access arrives at the old owner before
the relocation.

\subsection{Consistency}
\label{sec:consistency}

In this section, we analyze the consistency properties of \sys{} and compare
them to classic PSs, i.e., to PS-Lite~\cite{pslite}. Table~\ref{tab:consistency}
shows a summary. Consistency guarantees affect
the convergence of ML algorithms in the distributed setting; in particular,
relaxed consistency can slow down convergence~\cite{ssp,essp}. The extent of
this impact differs from task to task~\cite{ssp}. None of the existing PSs
guarantee serializability, as pull and push operations of different workers can
overlap arbitrarily. Neither do PSs give guarantees across multiple keys. PSs
can, however, provide per-key sequential consistency. Sequential consistency
provides two properties~\cite{sequentialconsistency}: (1) each worker's
operations are executed in the order specified by the worker, and (2) the result
of any execution is equivalent to an execution of the operations of all workers
in some sequential order. In the following, we study per-key sequential
consistency for synchronous and asynchronous operations. Note that stale PSs do
not provide sequential consistency, as discussed in Section~\ref{sec:ps}. We
assume in the following that nodes process messages in the order they arrive
(which is true for PS-Lite and \sys{}).

\textbf{Synchronous operations.} A classic PS guarantees sequential consistency:
it provides property (1) because workers block during synchronous operations,
preventing reordering, and (2) because all operations on one parameter are
performed sequentially by its owner.

\vspace{-0.1cm}
\begin{theorem}
  \label{theorem:sync}
  \sys{} guarantees sequential consistency for synchronous operations.
\end{theorem}
\proof{} In the absence of relocations, \sys{} provides sequential
consistency, analogously to classic PSs. In the presence of relocations, it
provides property (1) because synchronous operations also block the worker if a
parameter relocates. It provides property (2) because, at each point in time,
only one node processes operations for one parameter. During a relocation, the
old owner processes operations until the parameter leaves the local store
(at this time, no further operations for this key are in the old owner's queue). Then
the parameter is transferred to the new owner, which then starts processing. The
new owner queues concurrent operations until the relocation is finished. \sys{}
employs latches to guarantee a sequential execution among local threads.

\textbf{Asynchronous operations.} A classic PS such as PS-Lite provides
sequential consistency for asynchronous operations.\footnote{We assume that the
  network layer preserves message order. This is the case in PS-Lite and \sys{}
  because they use TCP and send operations of a thread over the
  same connection.} Property (1) requires that operations reach the responsible
server in program order (as the worker does not block during an asynchronous
operation). This is the case in PS-Lite as it sends each message directly to the
responsible server. Property (2) is given as for synchronous operations.

\vspace{-0.1cm}
\begin{theorem}
  \sys{} without location caches guarantees sequential consistency for
  asynchronous operations.
\end{theorem}
\proof{} For property (1), first suppose that there is no concurrent relocation.
\sys{} routes the operations of a worker on one parameter to the parameter's
home node and from there to the owner. Message order is preserved in both steps
under our assumptions. Now suppose that the parameter is relocated in-between
operations. In this case, the old owner processes all operations that arrive at
the home node before the relocation. Then the parameter is moved to the new
owner, which then takes over and processes all operations that arrive at the
home node after the relocation. Again, message order is preserved in all steps,
such that \sys{} provides property (1). It provides property (2) by the same argument as
for Theorem~\ref{theorem:sync}.

\vspace{-0.1cm}
\begin{theorem}
  \sys{} with location caches does not provide sequential consistency for
  asynchronous operations.
\end{theorem}
\proof{} \sys{} does not provide property (1) because a location cache change
can cause two operations to be routed differently, which can change message
order at the recipient. For example, consider two operations $O_1$ and $O_2$. $O_1$ is sent to
the currently cached, but outdated, owner. Then the location cache is updated
(by another returning operation) and $O_2$ is sent directly to the current
owner. Now, it is possible that $O_2$ is processed before $O_1$, because $O_1$
has to be double-forwarded to the current owner. This breaks sequential, causal,
and PRAM consistency.

\subsection{Location Management}
\label{sec:management}

\begin{table}
  \caption{Location management strategies. $N$ is the number of nodes, $K$ is the
    number of parameter keys.}
  \label{tab:ownership-management}
  \centering
  \scriptsize
\begin{threeparttable}
\begin{tabular}{lccc}
\toprule
  Strategy & Storage & \multicolumn{2}{c}{Number of messages for} \\
  \cmidrule(lr){2-2} \cmidrule(lr){3-4}
                               & (per node) &   remote access & relocation\\
\midrule
Static partition & 0 & \(2\) & n/a\\
Broadcast operations & 0 & \(N\) & \(0\)\\
Broadcast relocations & \(K\) & \(2\) & \(N\)\\
Home node & \(K/N\) & \(3\)\tnote{a} & \(3\)\\
\bottomrule
\end{tabular}
\begin{tablenotes}
\item[a] 3 messages if uncached, 2 with correct cache, 4 with stale cache
\end{tablenotes}
\end{threeparttable}
\figurespace{}
\end{table}


\begin{table*}
  \caption{ML tasks, models, and datasets.
    The rightmost columns depict the number of key
    accesses and the size of read parameters (per second, for a single thread),
    respectively. }
  \label{tab:datasets}
  \centering
  \scriptsize
  \begin{threeparttable}
  \begin{tabular}{llrrrlrrrr} \toprule
      Task & \multicolumn{4}{c}{Model parameters} & \multicolumn{3}{c}{Data} & \multicolumn{2}{c}{Param.~Access}                                                    \\
    \cmidrule(lr){2-5} \cmidrule(lr){6-8} \cmidrule(lr){9-10}
           & Model                     & \mbox{Keys}           & \mbox{Values} & Size & Data set & \mbox{\hspace{-2cm}Data points} & Size & Keys/s & MB/s \\ \midrule

    Matrix & \multicolumn{1}{l}{Latent Factors, rank~100} & \SI{6.4}{M} & \SI{640}{M} & \SI{4.8}{GB} & \sqM{} matrix & \SI{1000}{M} & \SI{31}{GB} & \SI{414}{k} & \num{315} \\
    \ \ Factorization            & \multicolumn{1}{l}{Latent Factors, rank~100} & \SI{11.0}{M}& \SI{1100}{M} & \SI{8.2}{GB} & \rectM{} matrix & \SI{1000}{M} & \SI{31}{GB} & \SI{316}{k} & \num{241} \\

    \addlinespace[0.4em]
    Knowledge & \multicolumn{1}{l}{ComplEx, dim.~\num{100}} & \SI{0.5}{M} & \SI{98}{M} & \SI{0.7}{GB} & DBpedia-500k & \SI{3}{M} & \SI{47}{MB} & \SI{312}{k}  & \num{476} \\
    \ \  Graph          & \multicolumn{1}{l}{ComplEx, dim.~\num{4000}} & \SI{0.5}{M} & \SI{3929}{M} & \SI{29.3}{GB} & DBpedia-500k & \SI{3}{M} & \SI{47}{MB} & \SI{11}{k} & \num{643} \\
    \ \ Embeddings          & \multicolumn{1}{l}{RESCAL, dim.~\num{100}} & \SI{0.5}{M} & \SI{110}{M} & \SI{0.8}{GB} & DBpedia-500k & \SI{3}{M} & \SI{47}{MB} & \SI{12}{k} & \num{614} \\

    \addlinespace[0.4em]
    Word Vectors & \multicolumn{1}{l}{Word2Vec, dim.~\num{1000}} & \SI{1.1}{M} & \SI{1102}{M} & \SI{4.1}{GB} & 1b word benchmark & \SI{375}{M} & \SI{3}{GB} & \SI{17}{k} & \num{65} \\
    \bottomrule
  \end{tabular}
  \end{threeparttable}
  \figurespace{}
\end{table*}


There are several strategies for managing location information in a PS with DPA.
Key questions are how to store and communicate knowledge about which server is
currently responsible for a parameter. Table~\ref{tab:ownership-management}
contrasts several possible strategies. For reference, we include the static
partitioning of existing PSs (which does not support DPA). In the following, we
discuss the different strategies. We refer to the number of nodes as $N$ and to
the number of keys as~$K$.

\textbf{Broadcast operations. } One strategy is to avoid storing any location
information and instead broadcast the request to all nodes for each non-local
parameter access. Then, only the server that currently holds the parameter
responds to the request (all other servers ignore the message). This requires no
storage but sends $N$ messages per parameter access ($N-1$ messages to all other
nodes, one reply back to the requester). This high communication cost is not
acceptable within a PS.

\textbf{Broadcast relocations. } An alternative strategy is to replicate location
information to all nodes. This requires to store $K$~locations on each node (one
for each of $K$ keys). An advantage of this approach is that only two messages
are required per remote parameter access (one request to the current owner of
the parameter and the response). However, storage cost may be high when there is
a large number of parameters and each location change has to be propagated
to all nodes. The simplest way to do this is via direct mail, i.e., by sending
$N-2$ additional messages to inform all nodes that were not involved in a
parameter relocation. Gossip protocols~\cite{gossip} could reduce this
communication overhead.

\textbf{Home node.} \sys{} uses a home node strategy, inspired by distributed
hash tables~\cite{can, chord, pastry, tapestry} and home-based approaches in
general~\cite{steen17}. The home node of a parameter knows which node currently
holds the parameter. Thus, if any node does not know the location of a
parameter, it sends a request to the home node of that parameter. As discussed
in Section~\ref{sec:access}, this requires at least one additional message for
remote parameter access. A home node is assigned to each parameter using static
partitioning, e.g., using range or hash partitioning. A simpler, but not
scalable variant of this strategy is to have a centralized home node that knows
the locations of all parameters. We discard this strategy because it limits the
number of parameters to the size of one node and creates a bottleneck at the
central home node.

\sys{} employs the (decentralized) home node strategy because it requires little
storage overhead and sends few messages for remote parameter access, especially
when paired with location caches.

\subsection{Granularity of Location Management}
\label{sec:granularity}

Location can be managed at different granularities, e.g., for each key or for
ranges of keys. \sys{} manages parameter location per key and allows
applications to localize multiple parameters in a single localize operation.

This provides high flexibility, but can cause overhead if applications do not
require fine-grained location control. For example, parameter blocking
algorithms~\cite{dsgd,dsgdpp,nomad} relocate parameters exclusively in static
(pre-defined) blocks. For such algorithms, a possible optimization would be to
manage location on group level. This would reduce storage requirements and would
allow the system to optimize for communication of these groups. We do not
consider such optimizations because \sys{} aims to support many PAL methods,
including ones that require fine-grained location control, such as latency
hiding.

\subsection{Important Implementation Aspects}
\label{sec:implementation}

In this section, we discuss implementation aspects that are key
for the performance or the consistency of \sys{}.

\textbf{Message grouping.} If a single push, pull, or localize operation
  includes more than one parameter, \sys{} groups messages that go to the same
  node to reduce network overhead. For example, consider that one localize call
  localizes multiple parameters. If two of the parameters are managed by the
  same home node, \sys{} sends only one message from requester to this home
  node. If the two parameters then also currently reside at the same location,
  \sys{} again sends only one message from home node to the current owner and
  one back from the current owner to the requester. Message grouping adds system
  complexity, but is very beneficial when clients access or localize sets of
  parameters at once.

\textbf{Local parameter store.} As other PSs~\cite{ssp,pslite}, \sys{} provides
two variants for the local parameter store: dense arrays and sparse maps. Dense
parameter storage is suitable if parameter keys are contiguous; sparse storage
is suitable when they are not. \sys{} uses a list of $L$ latches to synchronize
parameter access, while allowing parallel access to different parameters. A
parameter with key $k$ is protected by latch $k \operatorname{mod} L$. Applications can
customize $L$. A default value of $L=1000$ latches worked well in our experiments.

\textbf{No message prioritization.} To reduce blocking time, \sys{} could have
opted to prioritize the processing of messages that belong to parameter
relocations. However, this prioritization would break most consistency guarantees
for asynchronous operations (i.e., sequential, causal, and PRAM
consistency). The reason for this is that an ``instruct relocation'' message could
overtake a parameter access message at the old owner of a relocation. The old owner
would then reroute the parameter access message, such that it potentially
arrives at the new owner after parameter access messages that were issued later.
Therefore, \sys{} does not prioritize messages.


\section{Experiments}
\label{sec:experiments}

We conducted an experimental study to investigate the efficiency of classic PSs
(Section~\ref{sec:exp:classic}) and whether it is beneficial to integrate PAL
techniques into PSs (Section~\ref{sec:exp:dynamic}). Further, we investigated
how efficient \sys{} is in comparison to a task-specific low-level
implementation (Section~\ref{sec:exp:mpi}) and stale PSs
(Section~\ref{sec:exp:stale}), and conducted an ablation study
  (Section~\ref{sec:exp:ablation}). Our major insights are: (i) Classic PSs
suffered from severe communication overhead compared to a single node: using
Classic PSs, 2--8 nodes were slower than 1~node in all tested tasks. (ii)
Integrating PAL techniques into the PS reduced this communication overhead:
\sys{} was 4--203x faster than a classic PS, with 8 nodes outperforming
1~node by up to 9x. (iii) \sys{} scaled better than a state-of-the-art stale PS
(8 nodes were 9x vs.~2.9x faster than 1 node).

\subsection{Experimental Setup}
\label{sec:experiments:setup}

We considered three popular ML tasks that require long training: matrix
factorization, knowledge graph embeddings, and word vectors.
Table~\ref{tab:datasets} summarizes details about the models and the datasets that we
used for these tasks. We employ varied PAL techniques for the tasks. In the
following, we briefly discuss each task.
\iflong
Appendix~\ref{sec:task-details} provides further details.
\else
Appendix~A of a long version of this paper~\cite{lapse-long} provides further details.
\fi

\textbf{Matrix factorization. } Low-rank matrix factorization is a common tool
for analyzing and modeling dyadic data, e.g., in collaborative filtering for
recommender systems~\cite{mf-recsys}. We employed a parameter blocking
approach~\cite{dsgd} to create and exploit PAL: communication happens only
between subepochs; within a subepoch, all parameter access is local. We
implemented this algorithm in PS-Lite (a classic PS), Petuum (a stale PS), and
\sys{}. Further, we compared to a task-specific and tuned low-level
implementation of this parameter blocking
approach.\footnote{\url{https://github.com/uma-pi1/DSGDpp}} We used two synthetic
datasets from \cite{mf-kais}, because the largest openly available dataset that
we are aware of is only \SI{7.6}{GB} large.

\textbf{Knowledge graph embeddings. } Knowledge graph embedding (KGE) models
learn algebraic representations of the entities and relations in a knowledge
graph. For example, these representations have been applied successfully to
infer missing links in knowledge graphs~\cite{kgcompletion}. A vast number of
KGE models has been proposed~\cite{rescal,transe,hole,distmult,analogy}, with
different training techniques~\cite{kgetraining}. We studied two models as
representatives: RESCAL~\cite{rescal} and ComplEx~\cite{complex}. We employed data
clustering and latency hiding to create and exploit PAL. We used
the DBpedia-500k dataset~\cite{dbpedia500}, a real-world knowledge graph that
contains \num{490598} entities and \num{573} relations of
DBpedia~\cite{dbpedia}.

\textbf{Word vectors. } Word vectors are a language modeling technique in
natural language processing: each word of a vocabulary is mapped to a vector of
real numbers~\cite{word2vec,glove,elmo}. These vectors are useful as input for
many natural language processing tasks, for example, syntactic
parsing~\cite{syntactic-parsing} or question
answering~\cite{question-answering}. In our experimental study, we used the
skip-gram Word2Vec~\cite{word2vec} model and employed latency hiding to create
and exploit PAL. We used the One Billion Word Benchmark~\cite{1b-word} dataset,
with stop words of the Gensim~\cite{gensim} stop word list removed.

\textbf{Implementation and cluster.} We implemented \sys{} in C++, using ZeroMQ
and Protocol Buffers for communication, drawing from PS-Lite~\cite{pslite}. Our
code is open source and we provide information on how to reproduce our
experiments.\footnote{\url{https://github.com/alexrenz/lapse-ps/}} We ran version~1.1 of Petuum\footnote{In consultation with
  the Petuum authors, we fixed an issue in Petuum that prevented Petuum from
  running large models on a single node.}~\cite{petuum} and the
version of Sep~1, 2019 of PS-Lite~\cite{pslite}. We used a local cluster of 8 Dell
PowerEdge R720 computers, running CentOS Linux 7.6.1810, connected with
10~GBit Ethernet. Each node was equipped with two Intel Xeon E5-2640 v2 8-core
CPUs, 128~GB of main memory, and four 2~TB NL-SAS 7200~RPM hard disks. We
compiled all code with gcc 4.8.5.

\begin{figure}[t]
  \includegraphics[page=1,width=1\columnwidth]{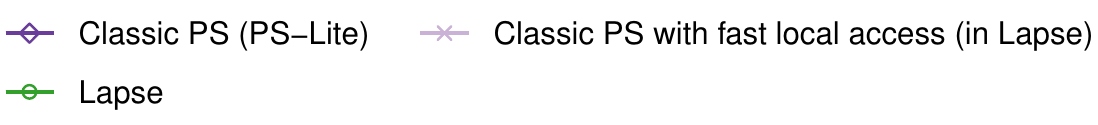}
  \begin{subfigure}[b]{.235\textwidth}
    \centering
    \includegraphics[page=1,width=1.0\textwidth]{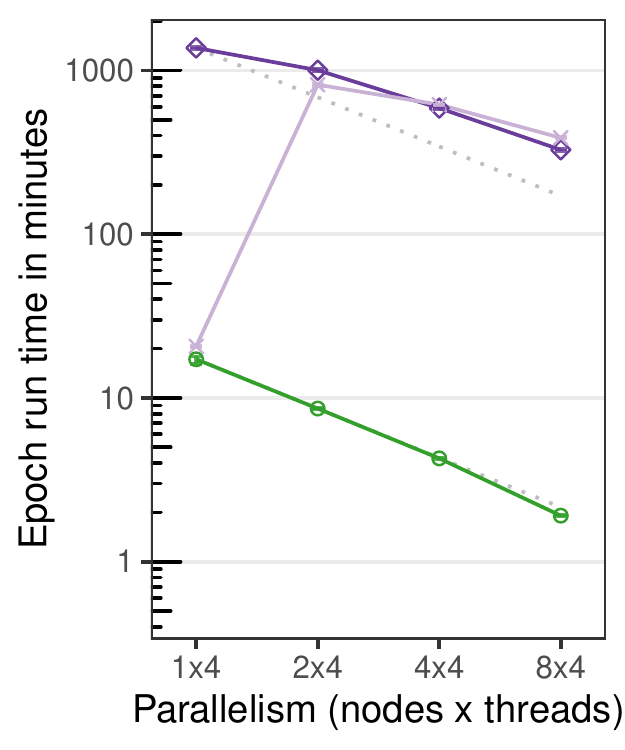}
    \caption{\rectM{} matrix, 1b entries}
    \label{fig:mf:rect}
  \end{subfigure}
  \begin{subfigure}[b]{0.235\textwidth}
    \centering
    \includegraphics[page=1,width=1.0\textwidth]{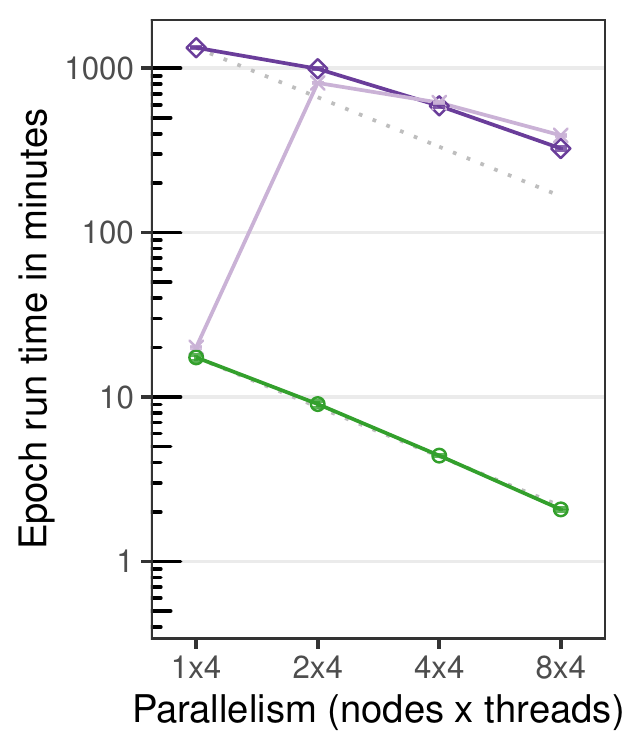}
    \caption{\sqM{} matrix, 1b entries}
    \label{fig:mf:sq}
  \end{subfigure}
  \caption{Performance results for matrix factorization. \sys{} scaled
    above-linearly because it exploits PAL. Classic PS approaches displayed
    significant communication overhead over the single node. The classic PS
    approach in \sys{} drops in performance because it is efficient on a
    single node, see Section~\ref{sec:exp:classic}. The gray dotted lines
    indicate linear scaling. Error bars depict minimum and maximum run
    time (hardly visible here because of low variance).}
  \label{fig:mf}
  \figurespace{}
\end{figure}


\textbf{Settings and measures.} In all experiments, we used 1~server and 4
worker threads per node and stored all model parameters in the PS, using dense
storage.
Each key held a vector of parameter values. 
We report \sys{} run times without location caches, because they
  had minimal effect in \sys{}. The reason for this is that \sys{} localizes
  parameters and location caches are not beneficial for local parameters (see Section~\ref{sec:exp:ablation} for details).
For all tasks but word vectors, we measured epoch run time, because epochs are
identical (or near-identical). This allowed us to conduct experiments in more
reasonable time. For word vectors, epochs are not identical because the chosen
latency hiding approach changes the sampling distribution of negative samples
(%
\iflong
see Appendix~\ref{sec:task-details}%
\else
see Appendix~A of the long version~\cite{lapse-long}%
\fi%
). Thus, we measure model accuracy over time. We calculated model accuracy using
a common analogical reasoning task of \num{19544} semantic and syntactic
questions~\cite{word2vec}. We conducted 3 independent runs of each experiment
and report the mean. Error bars depict the minimum and maximum. In some
experiments, error bars are not clearly visible because of small variance. Gray
dotted lines indicate run times of linear scaling.

\begin{figure*}[t]
  \centering
  \includegraphics[page=1,width=0.82\textwidth]{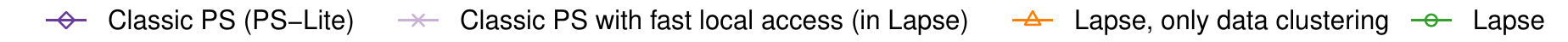}

  \begin{subfigure}[b]{.33\textwidth}
    \centering
    \includegraphics[page=1,width=0.9\textwidth]{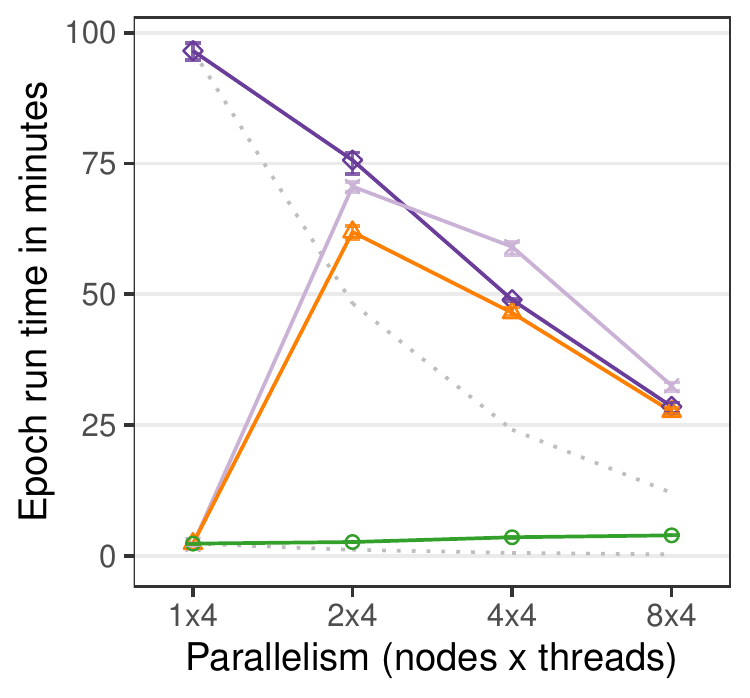}
    \caption{ComplEx-Small (dim.~\num{100}/\num{100})}
    \label{fig:kge:complex100}
  \end{subfigure}%
  \begin{subfigure}[b]{.33\textwidth}
    \centering
    \includegraphics[page=1,width=0.9\textwidth]{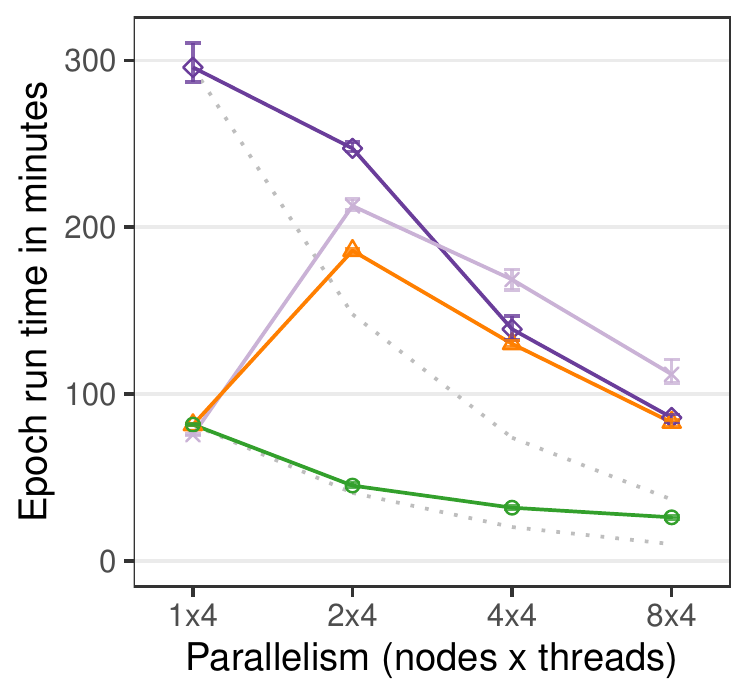}
    \caption{ComplEx-Large (dim.~\num{4000}/\num{4000})}
    \label{fig:kge:complex4000}
  \end{subfigure}%
  \begin{subfigure}[b]{.33\textwidth}
    \centering
    \includegraphics[page=1,width=0.9\textwidth]{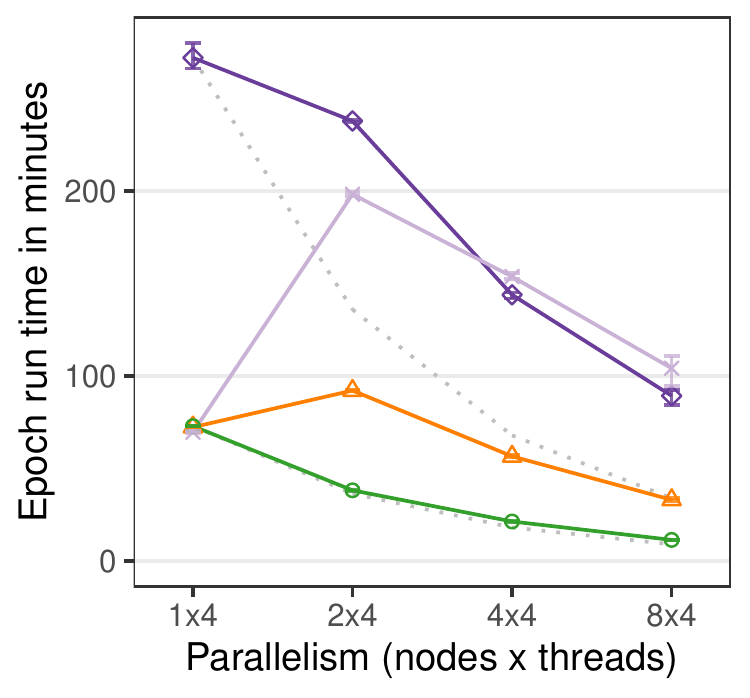}
    \caption{RESCAL-Large (dim.~\num{100}/\num{10000})}
    \label{fig:kge:rescal100}
  \end{subfigure}%
  \caption{Performance results for training knowledge graph embeddings.
    Distributed training using the classic PS approach did not outperform a
    single node in any task. \sys{} scaled well for the large tasks (b and c),
    but not for the small task (a).}
  \label{fig:kge}
  \figurespace{}
\end{figure*}


\subsection{Performance of Classic Parameter Servers}
\label{sec:exp:classic}
We investigated the performance of classic PSs and how it compares to the
performance of efficient single-machine implementations. To this end, we
measured the performance of a classic PS on 1--8 nodes for matrix factorization
(Figure~\ref{fig:mf}), knowledge graph embeddings (Figure~\ref{fig:kge}), and
word vectors (Figure~\ref{fig:we}). Besides PS-Lite, we ran \sys{} as a
  classic PS (with shared memory access to local parameters). To do so, we
  disabled DPA, such that parameters are allocated statically. We used random
  keys for the parameters in both implementations.\footnote{The performance of
    classic PSs depends on the (static) assignment of parameters. Both
    implementations range partition parameters, which can be suboptimal if
    algorithms assign keys to parameters non-randomly. Manually assigning random
    keys improved performance for most tasks (and never deteriorated
    performance).} We omitted PS-Lite from the word vector task due to its run
time.

\textbf{Multi-node performance.} The performance of classic PSs was dominated by
communication overhead: in none of the tested ML tasks did 2--8 nodes outperform
a single node. Instead, 2--8 nodes were 22--47x slower than 1~node for matrix
factorization, 1.4--30x slower for knowledge graph embeddings, and 11x slower
for word vectors. The two classic PS implementations displayed similar
performance on multiple nodes. With smaller numbers of nodes (e.g., on 2 nodes),
the variant with fast local access can access a larger part of parameters
with low latency, and thus has a performance benefit. Further performance differences
stem from differences in the system implementations.

\textbf{Single-node performance.} On a single node, the run times of PS-Lite and
\sys{} differed significantly (e.g., see Figure~\ref{fig:mf}), because they access local
parameters (i.e., all parameters when using 1~node) differently. \sys{} accesses
local parameters via shared memory. This was 71--91x faster than PS-Lite, which
accesses local parameters via inter-process communication.\footnote{ PS-Lite
  provides an option to explicitly speed up single node performance by using
  memory copy inter-process communication. In our experiments, this was still
  47--61x slower than shared memory.}
The Classic PS with fast local access displayed the same single-node performance
as \sys{}, as all parameters are local if only one node is used (even without
relocations). This single-node efficiency is the reason for its performance drop
from 1 to 2~nodes. Comparing distributed run times only against inefficient
single-node implementations can be misleading.

\textbf{Communication overhead.} The extent of communication overhead depended
on the task-specific communication-to-computation ratio. The two rightmost columns
of Table~\ref{tab:datasets} give an indication of this ratio. They depict the
number of key accesses and size of read parameter data per second, respectively,
measured for a single thread on a single node for the respective task. For
example, ComplEx-Small (Figure~\ref{fig:kge:complex100}) accessed the PS
frequently (312k accessed keys per second) and displayed high communication
overhead (8~nodes were 14x slower than 1~node). ComplEx-Large
(Figure~\ref{fig:kge:complex4000}) accessed the PS less frequently (11k
accesses keys per second) and displayed lower communication overhead (8~nodes were
1.4x slower than 1~node).

\begin{table}
  \caption{Parameter reads, relocations, and relocation times in ComplEx-Large.
    In this task, each key holds a vector of \num{8000} doubles. All
    parallelism levels read \num{196} million keys in one epoch. On 2 nodes,
    mean RT is short as every relocation involves only 2 nodes (instead of~3).}
  \label{tab:reloc-stats}
  \centering
  \scriptsize
  \begin{threeparttable}
    \begin{tabular}{crrrrr}
      \toprule
      Nodes & \multicolumn{3}{c}{Reads (keys/s)} & Relocations & Mean RT\tnote{a}\\
      \cmidrule(lr){2-4} \cmidrule(lr){5-5} \cmidrule(lr){6-6}
      & Total & Local & Non-local & (keys/s) & (ms) \\
      \midrule
      1 & \SI{36}{k} & \SI{36}{k} & \SI{0.0}{k} & \SI{0}{k} & -\\ 
      2 & \SI{72}{k} & \SI{72}{k} & \SI{0.0}{k} & \SI{12}{k} & \num{2.4}\\ 
      4 & \SI{104}{k} & \SI{102}{k} & \SI{1.6}{k} & \SI{27}{k} & \num{6.9}\\ 
      8 & \SI{121}{k} & \SI{118}{k} & \SI{2.5}{k} & \SI{36}{k} & \num{7.7}\\ 
      \bottomrule
    \end{tabular}
    \begin{tablenotes}
    \item[a] Relocation time, see Section~\ref{sec:movement}
    \end{tablenotes}
  \end{threeparttable}
  \figurespace{}
\end{table}


\subsection{Effect of Dynamic Parameter Allocation}
\label{sec:exp:dynamic}

\begin{figure*}[t]
  \centering
  \includegraphics[page=1,width=0.9\textwidth]{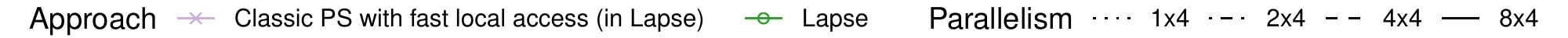}

  \begin{subfigure}[b]{.33\textwidth}
    \centering
    \includegraphics[page=1,width=0.9\textwidth]{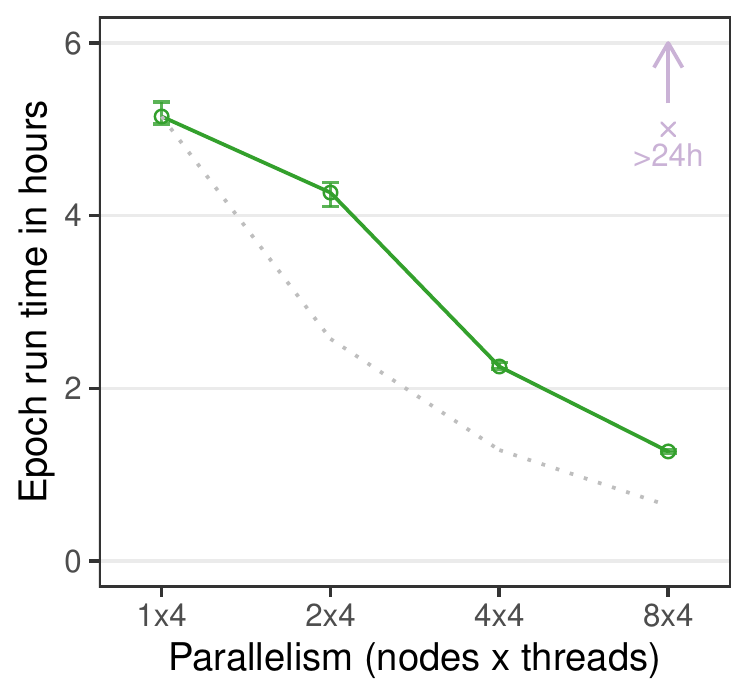}
    \caption{Epoch runtime}
    \label{fig:we:epochtime}
  \end{subfigure}%
  \begin{subfigure}[b]{.33\textwidth}
    \centering
    \includegraphics[page=1,width=0.9\textwidth]{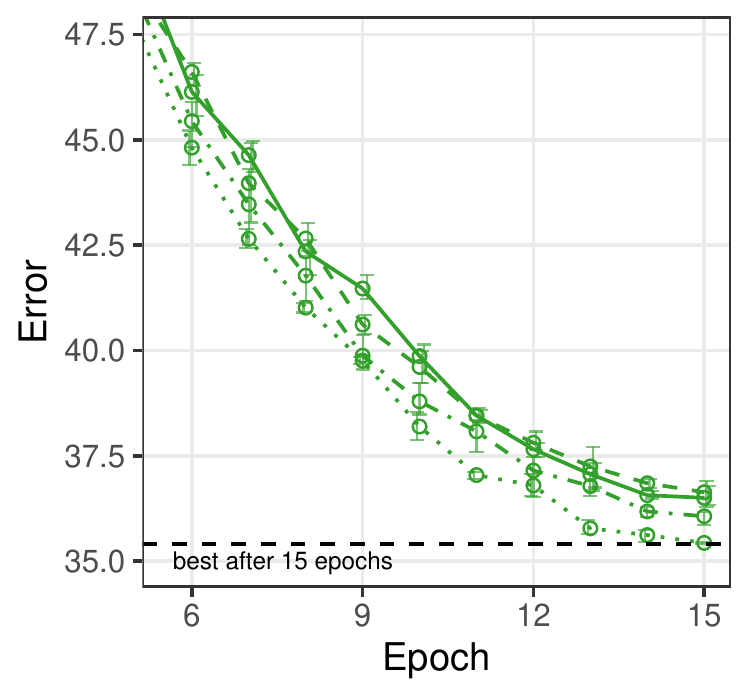}
    \caption{Error over epochs 6--15}
    \label{fig:we:epoch}
  \end{subfigure}%
  \begin{subfigure}[b]{.33\textwidth}
    \centering
    \includegraphics[page=1,width=0.9\textwidth]{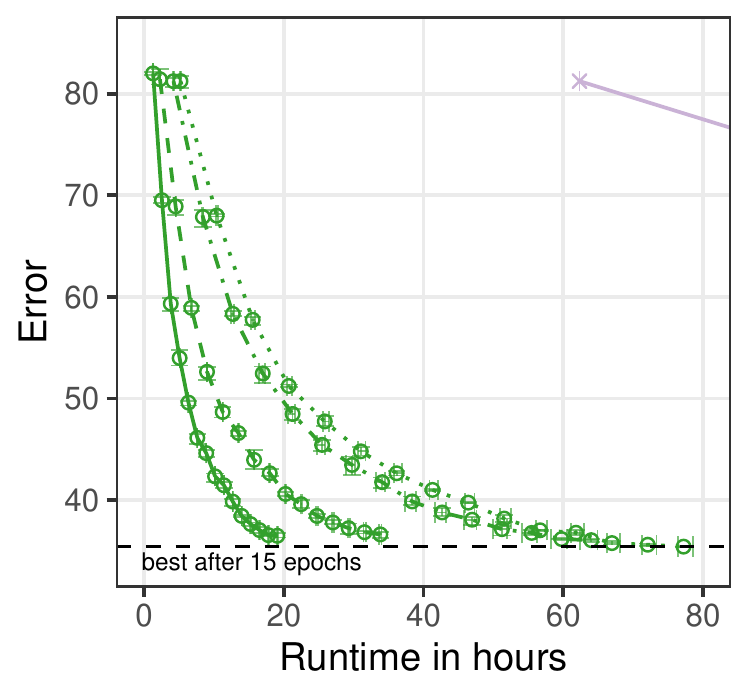}
    \caption{Error over run time (all epochs)}
    \label{fig:we:time}
  \end{subfigure}%
  \caption{Performance results for training word vectors. The classic PS
    approach did not scale (8 nodes were \textgreater4x slower than 1 node). In
    \sys{}, 8 nodes reached (for example) 39\% error 3.9x faster than 1~node.
    The dashed horizontal line indicates the best observed error after 15~epochs.}
  \label{fig:we}
  \figurespace{}
  \vspace{-0.2cm} 
\end{figure*}

We compared the performance of \sys{} to a classic PS
approach for matrix factorization (Figure~\ref{fig:mf}), KGE
(Figure~\ref{fig:kge}), and word vectors (Figure~\ref{fig:we}).
\sys{} was 4--203x faster than classic PSs. \sys{} outperformed the single node
in all but one tasks (ComplEx-Small), with speed-ups
of 3.1--9x on 8 nodes (over 1~node).

\textbf{Matrix factorization. } In matrix factorization, \sys{} was 90--203x
faster than classic PSs and achieved linear speed-ups over the single node (see
Figure~\ref{fig:mf}). The reason for this speed-up is that classic PSs (e.g.,
PS-Lite) cannot exploit the PAL of the parameter blocking algorithm. Thus, their
run time was dominated by network latency.

\textbf{Knowledge graph embeddings. } In knowledge graph embeddings, \sys{} was
4--26x faster than a classic PS (see Figure~\ref{fig:kge}). It scaled well for
the large tasks (ComplEx-Large and RESCAL-Large) despite localization conflicts
on frequently accessed parameters. The probability of a localization
  conflict, i.e., that two or more nodes localize the same parameter at the same
  time, increases with the number of workers, see Table~\ref{tab:reloc-stats}.
  In the table, the number of localization conflicts is indicated by the number
  of non-local parameter reads (which are caused by localization conflicts). For ComplEx-Small,
distributed execution in \sys{} did not outperform the single node because of
communication overhead. We additionally measured performance of running \sys{}
with only data clustering (i.e., without latency hiding). This approach accesses
relation parameters locally and entity parameters remotely. It improved
performance for RESCAL (Figure~\ref{fig:kge:rescal100}) more than for ComplEx
(Figures~\ref{fig:kge:complex100} and \ref{fig:kge:complex4000}), because in
RESCAL, relation embeddings have higher dimension (\num{10000} for RESCAL-Large)
than entity embeddings (\num{100}), whereas in ComplEx, both are the same size
(\num{100} in ComplEx-Small and \num{4000} in ComplEx-Large). 

\textbf{Word vectors. } For word vectors, \sys{} executed an epoch 44x faster
than a classic PS (Figure~\ref{fig:we}). Further, 8~nodes reached, for example,
39\% error 3.9x faster than a single node. The speed-up for word vectors is
lower than for knowledge graph embeddings, because word vector training exhibits
strongly skewed access to parameters: few parameters are accessed
frequently~\cite{word2vec}. This lead to more frequent localization conflicts in
the latency hiding approach than in knowledge graph embeddings, where negative
samples are sampled uniformly~\cite{kgetraining,analogy}.

\subsection{Comparison to Manual Management}
\label{sec:exp:mpi}

\begin{figure}[t]
  \includegraphics[page=1,width=1\columnwidth]{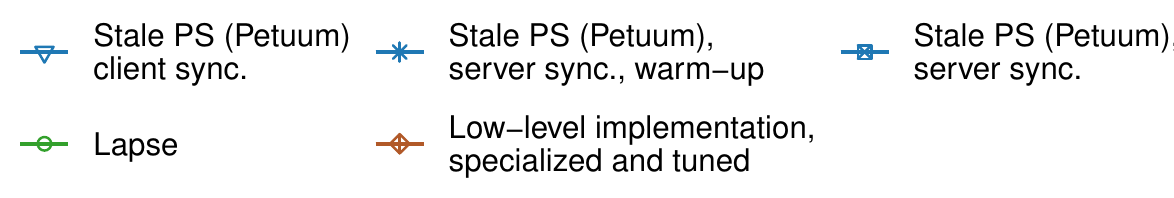}
  \begin{subfigure}[b]{.235\textwidth}
    \centering
    \includegraphics[page=1,width=1.0\textwidth]{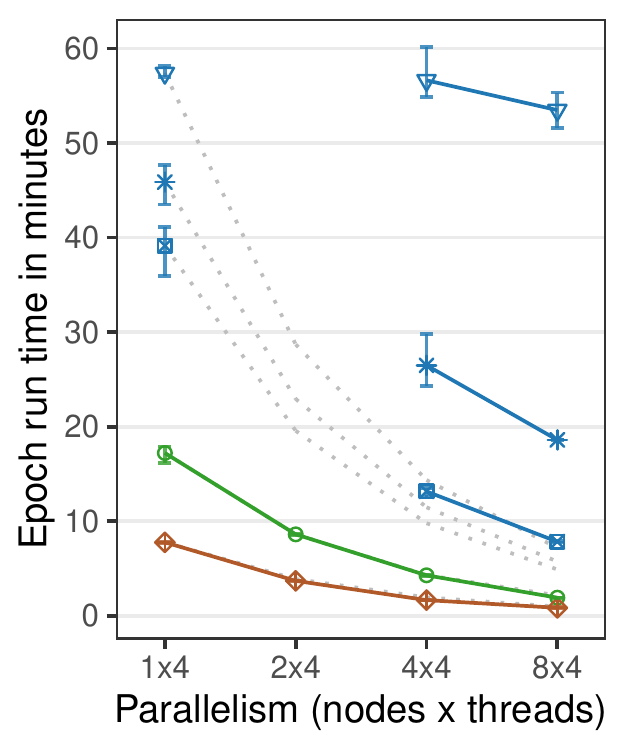}
    \caption{\rectM{} matrix, 1b entries}
    \label{fig:mf-other:rect}
  \end{subfigure}
  \begin{subfigure}[b]{0.235\textwidth}
    \centering
    \includegraphics[page=1,width=1.0\textwidth]{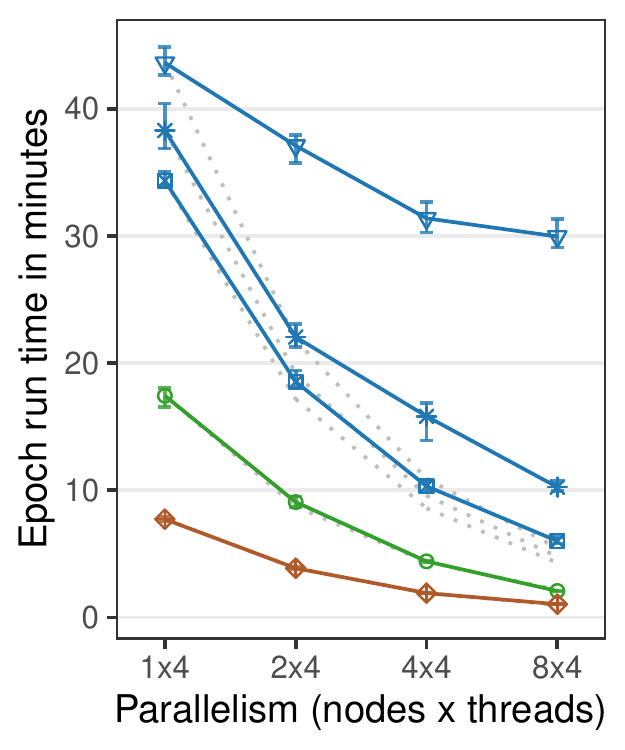}
    \caption{\sqM{} matrix, 1b entries}
    \label{fig:mf-other:sq}
  \end{subfigure}
  \caption{Performance comparison to manual parameter management and to Petuum,
    a state-of-the-art stale PS, for matrix factorization. \sys{} and manual
    parameter management (using a specialized and tuned low-level
    implementation) scale linearly, in contrast to Petuum. For the \rectM{}
    matrix, Petuum crashed with a network error on 2 nodes. }
  \label{fig:mf-other}
  \figurespace{}
\end{figure}


We compared the performance of \sys{} to a highly specialized and tuned
low-level implementation of the parameter blocking approach for matrix
factorization (see Figure~\ref{fig:mf-other}). This low-level
  implementation cannot be used for other ML tasks. Both the low-level
implementation and \sys{} scaled linearly (or slightly
above-linearly\footnote{The reason for the above-linear scaling for the
  \rectM{} matrix is that CPU caches work better the more workers are used,
  because each worker then focuses on a smaller part of the dataset and the
  model~\cite{dsgdpp}.}).

\sys{} had only 2.0--2.6x generalization overhead over the low-level
implementation. The reason for the overhead is that the low-level implementation
exploits task-specific properties that a PS cannot exploit in general if it aims
to provide PS consistency and isolation guarantees for a wide range of ML
tasks. I.e., the task-specific implementation lets workers work directly
on the data store, without copying data and without concurrency control. This
works for this particular algorithm, because each worker focuses on a separate
part of the model (at a time), but is not applicable in general. In contrast,
\sys{} and other PSs copy parameter data out of and back into the server,
causing overhead over the task-specific implementation. Additionally, the
low-level implementation focuses on and optimizes for communication of blocks of
parameters (which \sys{} does not) and does not use a key--value abstraction for
accessing keys. 

Implementing the parameter blocking approach was significantly easier in \sys{}
than using low-level programming. The low-level implementation manually moves
parameters from node to node, using MPI communication primitives. This manual
allocation required 100s of lines of MPI code. In contrast, in \sys{}, the same
allocation required only 4 lines of additional code.

\subsection{Comparison to Stale PSs}
\label{sec:exp:stale}

We compared \sys{} to Petuum, a popular stale PS, for matrix factorization (see
Figure~\ref{fig:mf-other}). We found that the stale PS was 2--28x slower than
\sys{} and did not scale linearly, in contrast to \sys{}.

Petuum provides bounded staleness consistency. As discussed in
Section~\ref{sec:blocking}, this can support synchronous parameter blocking
algorithms, such as the one we test for matrix factorization. We compared
separately to the two synchronization strategies that Petuum provides:
client-based and server-based. On the \rectM{} dataset, Petuum crashed on two
nodes with a network error.

\textbf{Client-based synchronization. } Client-based synchronization (\emph{SSP}
consistency model in Petuum) outperformed the classic PS, but was 2.5--28x
slower than \sys{}. The main reason for the overhead was network latency for
synchronizing parameters when their value became too stale. This approach did
not scale because the number of synchronizations per worker was constant
when increasing the number of workers (due to the increasing number of
subepochs).

\textbf{Server-based synchronization. } Server-based synchronization
(\emph{SSPPush} consistency model in Petuum) outperformed the classic PS, but
was 2--4x slower than \sys{}, and only 2.9x faster on 8 nodes than \sys{} on
1~node. The reason for this is that after every global clock advance, Petuum's
server-based synchronization eagerly synchronizes to a node all parameters that
this node accessed previously. On the one hand, this eliminated the network
latency overhead of client-based synchronization. On the other hand, this caused
significant unnecessary communication, causing the overhead over \sys{} and
preventing linear scale-out: in each subepoch, each node accesses only a subset
of all parameter blocks, but Petuum replicates all blocks. Petuum
``learns'' which parameters to replicate to which node in a slower
\emph{warm-up} epoch, depicted separately in Figure~\ref{fig:mf-other}.

\subsection{Ablation Study}
\label{sec:exp:ablation}
\textbf{DPA and fast local access. } \sys{} differs from classic PSs in two
  ways: (1) DPA and (2) shared memory access to local parameters. To investigate
  the effect of each difference separately, we compared the run time of three
  different variants: \emph{Classic PS (PS-Lite)} (neither DPA nor shared memory),
  \emph{Classic PS with fast local access (in \sys{})} (no DPA, but shared
  memory), and \emph{\sys{}} (DPA and shared memory). The run times of these variants
  can be compared in Figures~\ref{fig:mf} and \ref{fig:kge}. Without DPA, shared
  memory had limited effect, as many parameters were non-local and access times
  were thus dominated by network latency (except for the single-node case, in which
  all parameters are local). Combining DPA and shared memory yielded
  better performance: DPA ensures that parameters are local and shared memory
  ensures that access to local parameters is fast.


\textbf{Location caching. } All figures report run times of \sys{}
  \emph{without} location caching. We investigated the effect of location caching,
  which \sys{} supports optionally. We observed similar run times \emph{with}
  location caching. For example, for KGE (Figure~\ref{fig:kge}), \sys{} was
  max.~3\% faster and max.~2\% slower with location caching than without.
  The reason for this is that location caches speed up only remote parameter
  accesses (see Section~\ref{sec:access}). The latency hiding approach in KGE,
  however, localizes all parameters before they are used, such that the vast
  majority of parameter accesses are local (see
  Table~\ref{tab:reloc-stats}). For matrix factorization, location caching had
  no effect at all, because all parameter accesses were local (due to the
  parameter blocking approach). In the Classic PS variant of \sys{}, location
  caches had no effect because parameters remained at their home nodes
  throughout training (as they do in other classic PSs).


\section{Related Work}

We discuss related work on reducing PS communication and dynamic
allocation in general-purpose key--value stores.

\textbf{Dynamic parallelism}. FlexPS~\cite{flexps} reduces communication
overhead by executing different phases of an ML task with different levels of
parallelism and moving parameters to active nodes. However, it provides no
location control, moves parameters only between phases, and pauses the training
process for the move. This reduces communication overhead for some ML tasks, but
is not applicable to many others, e.g., the tasks that we consider in this
paper. FlexPS cannot be used for PAL techniques because it does not provide
fine-grained control over the location of parameters. \sys{}, in contrast, is
more general: it supports the FlexPS setting, but also provides fine-grained
location control and moves parameters without pausing workers.

\textbf{PS communication reduction.} We discuss replication and bounded
staleness in Section~\ref{sec:prelims}. Another approach to reducing
communication overhead in PSs is to combine reads and updates of multiple
workers locally before sending them to the remote server~\cite{adam, mxnet, ssp,
  flexps}. However, this technique may break sequential consistency and reduces
communication only if different workers access the same parameters at the same
time. Application-side communication reduction approaches include sending
lower-precision updates~\cite{seide14}, prioritizing important parts of
updates~\cite{jayarajan19}, applying filters to updates~\cite{li14nips}, and
increasing mini-batch size~\cite{goyal17fb}. Many of these approaches can be
combined with DPA, which makes for interesting areas of future work. To
  reduce the load on the PS, STRADS~\cite{strads} allows for parameter
  management through task-specific ring-based parameter passing schemes. These
  schemes can support a subset of parameter blocking approaches, but not data
  clustering or latency hiding. Developers manually implement such schemes as
  task-specific additions outside the PS. In contrast, \sys{} integrates PAL
  support into the PS framework.

\textbf{Dynamic allocation in key--value stores. }
DAL~\cite{dal}, a general-purpose key--value store, dynamically allocates a
data item at the node that accesses it most (after an adaptation period). In
theory, DAL could exploit data clustering and synchronous parameter blocking PAL
techniques, but no others (due to the adaptation period). However, DAL accesses
data items via inter-process communication, such that access latency is too high
to exploit PAL in ML algorithms. Husky~\cite{husky} allows applications to move
data items among nodes and provides fast access to local data items. However,
local data items can be accessed by only one worker. Thus, Husky can exploit
only PAL techniques in which each parameter is accessed by only one worker,
i.e., parameter blocking, but not data clustering or latency hiding.

\section{Conclusion}

We explored whether and to what extent PAL techniques---i.e., techniques that
distributed ML algorithms employ to reduce communication overhead---can be
supported in PSs, and whether such support is beneficial. To this end, we
introduced DPA to PSs and described \sys{}, an implementation of a PS with DPA.
We found that DPA can reduce the communication overhead of PSs significantly,
achieving up to linear scaling.

With these results, a next step is to facilitate the use of PAL methods in PSs.
For example, a possible direction is to research how PSs can automatically use
latency hiding through pre-localization. Another area for future work is to
improve the scalability of the latency hiding technique. The main bottleneck for
this technique were localization conflicts on frequently accessed parameters.
Alternative management approaches might be more suitable for these parameters.


\subsubsection*{Acknowledgments}
This work was supported by the German Ministry of
Education and Research (01IS17052) and the German Research Foundation (410830482). 

\iflong
\appendix

\section{Experimental Details}
\label{sec:task-details}

\textbf{Matrix factorization. } For both datasets, we ran a factorization of
rank 100. In all PSs, we ran a global barrier after each subepoch to ensure
consistency. In Petuum, to ensure consistent replicas, we issued one clock after
each subepoch and set a staleness threshold of 1. Petuum's own matrix
factorization implementation ran out of memory because it stores
dense matrices.

\textbf{Knowledge graph embeddings. } We ran the common setting of SGD with
AdaGrad~\cite{adagrad} and negative sampling~\cite{kgetraining, analogy}. We
stored the AdaGrad metadata in the PS. In all experiments, we generated
negative samples by perturbing both subject and object of positive triples 10
times. We set the initial learning rate for AdaGrad to \num{0.1}. We used data
clustering to create and exploit PAL for relation parameters, and latency hiding
for entity parameters. For the relation parameters, we partitioned the training
dataset by relation and allocated each relation parameter at the node that uses
it, such that all accesses to relation parameters are local. Regarding entity
parameters, each worker pre-localizes all parameters that it requires for the
data point that follows the current one, including the parameters for the
negative samples derived from this data point. The transfer of these parameters
then overlaps with the computation for the current data point. We tried looking
further into the future, e.g., localizing the parameters of a data point 2, 3,
10, or 100 data points into the future. We observed similar speed-ups for 2 and
3 and lower speed-ups for 10 and 100.

\textbf{Word vectors.} We used common model parameters~\cite{word2vec} of
embedding size \num{1000}, window size 5, minimum count~2, negative sampling
with 25 samples, and 1e-5 frequent word subsampling. We used a latency hiding approach that
pre-localizes parameters for the words of a sentence when it reads the sentence.
As negative samples, our approach chooses only parameters that are (currently)
available locally. This pool of locally available parameters changes constantly
as parameters are relocated when they occur in sentences. This approach changes
the local sampling distribution of negative examples at one node. However, it
mostly preserves the global sampling distribution, as each parameter is local at
exactly one node (except that frequent parameters are sampled
under-proportionately, because they relocate more often and are sampled nowhere
during a transfer).

\fi

\bibliographystyle{arwconf}
\bibliography{main}

\end{document}